%% file: main.tex
\documentclass[10pt,twocolumn,letterpaper]{article}

\usepackage{iccv}              


\definecolor{iccvblue}{rgb}{0.21,0.49,0.74}
\usepackage[pagebackref,breaklinks,colorlinks,allcolors=iccvblue]{hyperref}
\usepackage{natbib}
\usepackage{algpseudocode}
\usepackage{url}
\usepackage{colortbl}
\usepackage{multirow}
\usepackage{ulem} 
\usepackage{multicol}
\usepackage{indentfirst}
\usepackage{enumitem}
\usepackage[linesnumbered,ruled,vlined]{algorithm2e}
\usepackage{cancel}
\usepackage{booktabs}
\usepackage[font=small,labelfont=bf]{caption}
\usepackage{graphicx}
\usepackage{amssymb}
\usepackage{comment}
\usepackage{ulem}
\usepackage{makecell}
\usepackage{booktabs}       
\usepackage{siunitx}        
\usepackage{xcolor}         
\hypersetup{
    colorlinks=true,
    linkcolor=blue,   
    urlcolor=blue,    
    citecolor=blue    
}
\usepackage{setspace}

\usepackage{amsmath}

\newcommand\blfootnote[1]{%
  \begingroup
  \renewcommand\thefootnote{}\footnote{#1}%
  \addtocounter{footnote}{-1}%
  \endgroup
}

\renewcommand{\paragraph}[1]{\vspace{2pt}\noindent\textbf{#1.}}

\SetKwInput{KwInput}{Input}           
\SetKwInput{KwOutput}{Output}         
\def\paperID{3556} 
\def\confName{ICCV}
\def\confYear{2025}

\title{CIARD: Cyclic Iterative Adversarial Robustness Distillation}
\author{
    Liming Lu$^1$
    \quad 
    Shuchao Pang$^{1*}$
    \quad
    Xu Zheng$^{2,3*}$
    \quad
    Xiang Gu$^1$
    \quad
    \\
    Anan Du$^4$
    \quad
    Yunhuai Liu$^5$
    \quad
    Yongbin Zhou$^1$
    \\
    \\
    $^1$Nanjing University of Science and Technology, \\ $^2$HKUST(GZ), $^3$INSAIT, Sofia University, St. Kliment Ohridski, \\ $^4$Nanjing University of Industry Technology, $^5$Peking University\\
{\tt\small \{luliming,pangshuchao,eminentguxiang,zhouyongbin\}@njust.edu.cn} \\ \tt\small zhengxu128@gmail.com, anan.du@niit.edu.cn, yunhuai.liu@pku.edu.cn}

\begin{document}
\maketitle
\input{sec/0_abstract}

\input{sec/1_intro}

\input{sec/2_preliminary_background}
\input{sec/3_proposed_distillation_strategy}
\input{sec/4_experiments}

\input{sec/6_conclusion}
{
    \small
    \normalem
    \bibliographystyle{ieeenat_fullname}
    \bibliography{main}
}

\end{document}


\title{CIARD: Cyclic Iterative Adversarial Robustness Distillation}
\author{\authorBlock}
\maketitlesupplementary

\appendix

In the supplementary materials, we provide an extensive elaboration on additional experiments and results for the proposed CIARD. On one hand (Section~\ref{sectionA}), we present a more systematic description of CIARD's complete algorithmic workflow~\ref{algorithm_CIARD_Full}. On the other hand (Section~\ref{sectionB}), we have expanded our experimental scope by incorporating a wider range of models and attack methods for testing, thereby offering more comprehensive experimental results.

\section{Cyclic Iterative ARD} \label{sectionA}
This section provides a comprehensive description of the Cyclic Iterative Adversarial Robustness Distillation (CIARD) algorithm, including its adversarial example generation pipeline, multi-teacher knowledge transfer strategy, and dynamic optimization mechanisms. The detailed algorithm description of CIARD is outlined in Algorithm~\ref{algorithm_CIARD_Full}.

\begin{algorithm}[t]
\caption{Cyclic Iterative ARD (CIARD)}
\label{algorithm_CIARD_Full}
\SetAlgoLined
\KwIn{Clean teacher model $T_{\text{nat}}$, robust teacher model $T_{\text{adv}}$, student model $S(x|\theta_{s})$, clean images $x$ and labels $y$, perturbation bound $\Omega$, training epochs $T$, temperature $\tau$}
\KwIn{weights $w_{\text{nat}}=0.5$, $w_{\text{adv}}=0.5$, learning rate $\alpha=0.1$, teacher learning rate $\alpha_t=0.01$, weight learning rate $\eta=0.025$}

\For{$epoch = 1$ \KwTo $T$}{
    \For{each mini-batch $(x, y)$}{
        \textcolor[rgb]{1,0,0}{$//$ * Adversarial Example Generation * $//$}\\
        Generate adversarial examples via PGD:
        $x^* = \underset{\delta\in\Omega}{\arg\max}\ \text{CE}(S(x+\delta), y)$\;

        \textcolor[rgb]{1,0,0}{$//$ * Compute $\mathcal{L}_{\text{student}}$ * $//$}\\
        i)   Clean knowledge transfer:
        $\mathcal{L}_{\text{nat}} = \text{KL}(S(x), T_{\text{nat}}(x))$\;
        ii)  Robust knowledge transfer:
        $\mathcal{L}_{\text{adv}} = \text{KL}(S(x^*), T_{\text{adv}}(x^*))$\;
        iii) Push loss for robust specialization:
        $\mathcal{L}_{\text{push}} = \text{Push}(S(x^*), T_{\text{nat}}(x^*))$\;
        iv) Adaptive weight update:
        $\begin{aligned}
        \hat{\mathcal{L}}_{\text{nat}} &= \mathcal{L}_{\text{nat}}/\mathcal{L}_{\text{nat}}^{\text{init}}; \\
        \hat{\mathcal{L}}_{\text{adv}} &= \mathcal{L}_{\text{adv}}/\mathcal{L}_{\text{adv}}^{\text{init}}; \\
        w_{\text{nat}} &= w_{\text{nat}} - \eta(w_{\text{nat}} - \frac{\hat{\mathcal{L}}_{\text{nat}}}{\hat{\mathcal{L}}_{\text{nat}} + \hat{\mathcal{L}}_{\text{adv}}}); \\
        w_{\text{adv}} &= 1 - w_{\text{nat}};
        \end{aligned}$\\
        v)  Total student loss:
        $\mathcal{L}_{\text{student}} = w_{\text{adv}}\mathcal{L}_{\text{adv}} + w_{\text{nat}}\mathcal{L}_{\text{nat}} - \lambda\mathcal{L}_{\text{push}}$\;

        \textcolor[rgb]{1,0,0}{$//$ * Compute $\mathcal{L}_{\text{adv\_teacher}}$ * $//$}\\
        $\mathcal{L}_{\text{adv\_teacher}} = \text{CE}(T_{\text{adv}}(x^*), y)$\;
        
        Update student model $S$ using $\nabla_{\theta_S}\mathcal{L}_{\text{student}}$\;
        \If{$epoch > 50$}{
            Update $T_{\text{adv}}$ using $\nabla_{\theta_T}\mathcal{L}_{\text{adv\_teacher}}$\;
        }
    }
}
\end{algorithm}

\section{Supplementary Experiments}
\label{sectionB}

\noindent\textbf{Training Dynamics Analysis.}
Figure~\ref{fig:comparison} illustrates the evolution of both clean accuracy and robust accuracy throughout the training process. Unlike traditional ARD methods that typically show sharp trade-offs between these metrics, CIARD demonstrates a more harmonious progression where robust accuracy steadily improves without substantially sacrificing clean performance. Notably, after epoch 50 when the robust teacher begins updating, we observe an accelerated improvement in the student's adversarial robustness.

\begin{figure}[t!]
    \centering
    \includegraphics[width=1.0\linewidth]{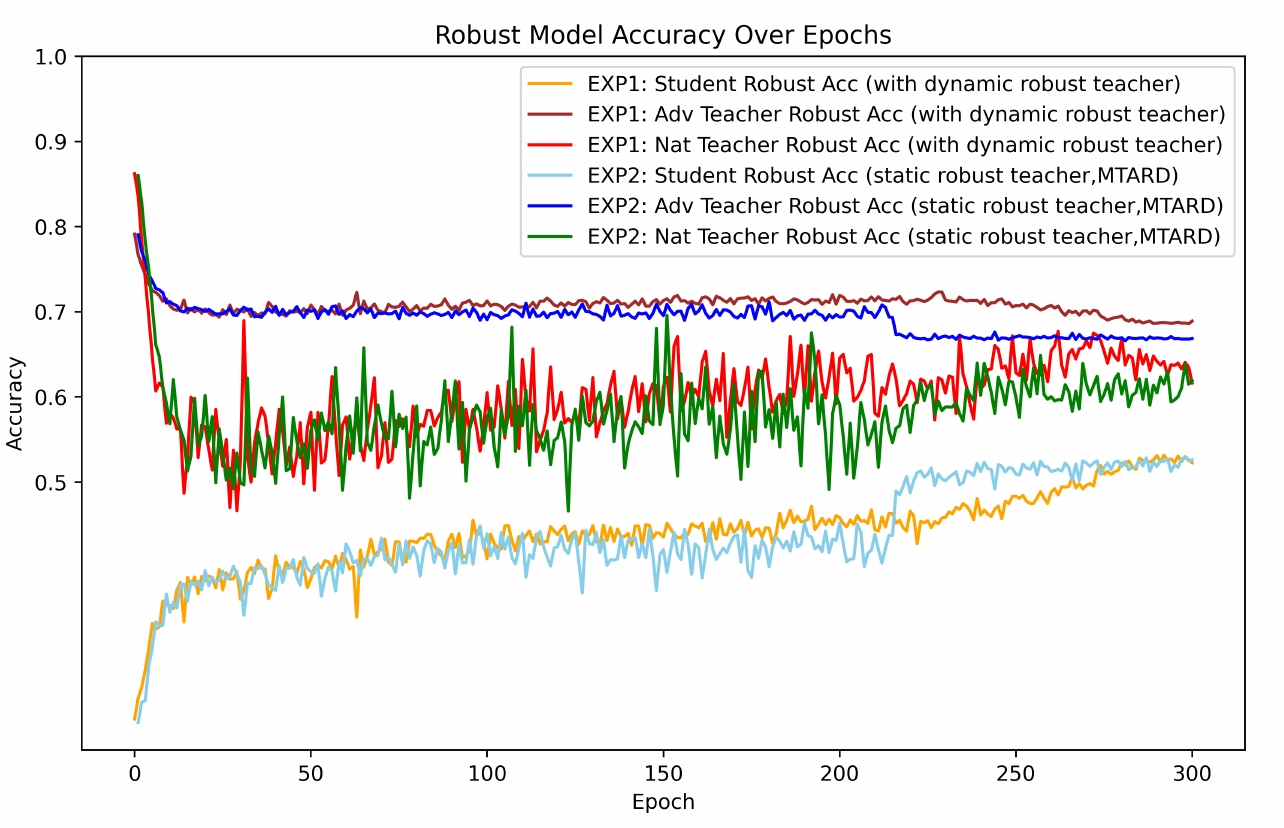}
    \vspace{-20pt}
    \caption{Robust Model Accuracy Over Epochs.}
    \label{fig:Robust_Model_Accuracy}
\end{figure}

In addition, Figure~\ref{fig:Robust_Model_Accuracy} shows the ablations study that only includes the learnable teacher and does not include the pushing loss component. The results clearly show that without the pushing loss, the robustness of our clean teacher decreases, and the robustness accuracy of the student model cannot be significantly improved. This highlights the crucial role of the pushing loss mechanism in our framework, which helps maintain the performance of the clean teacher while transferring effective robustness to the student model.

\begin{figure}[t!]
    \centering
    \includegraphics[width=1.0\linewidth]{Paper_Pictures/Picture4_Robust&Nature_y.pdf}
    \vspace{-20pt}
    \caption{MTARD vs. CIARD.}
    \label{fig:comparison}
\end{figure}

\noindent\textbf{Computational Efficiency Analysis.} While our Iterative Teacher Training (ITT) introduces additional computation, the overhead is moderate: when training the results in Table 3 (main paper) using WRN-34-10 as the robust teacher on a single RTX 4090 GPU, the per-epoch time increases from 111.06s (w/o ITT) to 140.56s (w/ ITT). This overhead largely depends on the teacher model size and remains relatively minor for lightweight architectures. We believe the trade-off is acceptable given the consistent robustness gains.

\noindent\textbf{Complete Results on CIFAR-10 \& CIFAR-100.}
In Tables~\ref{tab_white_box_cifar10_cifar100_ResNet-18_complete} and~\ref{tab_eval_White_cifar10_cifar100_MobileNet-V2} of the supplementary material, we present comprehensive evaluation results of our proposed CIARD method compared with state-of-the-art adversarial robustness approaches on the CIFAR-10 and CIFAR-100 datasets. We conduct experiments using ResNet-18 and MobileNet-V2 architectures against four different attack methods: FGSM, PGD\textsubscript{SAT}, PGD\textsubscript{TRADES}, and CW$_{\infty}$.

For the CIFAR-10 dataset, CIARD consistently outperforms all baseline methods across both architectures. With ResNet-18, CIARD achieves the highest clean accuracy (88.87\%) while maintaining superior robust accuracy under various attacks. For example, under FGSM attack, CIARD achieves 61.88\% robust accuracy and 75.38\% weighted robust accuracy, surpassing both single-teacher methods like RSLAD (60.41\%, 72.20\%) and dual-teacher approaches like B-MTARD (61.42\%, 74.81\%).

Similar performance advantages are observed with MobileNet-V2, where CIARD achieves 89.51\% clean accuracy and demonstrates consistent improvements in both robust and weighted robust metrics across all attack types. For instance, under FGSM attack, CIARD achieves 59.10\% robust accuracy and 74.31\% weighted robust accuracy, outperforming B-MTARD (58.79\%, 73.94\%).

For the CIFAR-100 dataset, CIARD also demonstrates superior performance. With ResNet-18 under FGSM attack, CIARD achieves 65.73\% clean accuracy, 34.47\% robust accuracy, and 50.10\% weighted robust accuracy, exceeding B-MTARD's performance (65.08\%, 34.21\%, 49.65\%). These results consistently hold across different attack types, with CIARD maintaining its advantage in the more challenging CIFAR-100 dataset.

The comprehensive experimental results confirm that our cyclic iterative approach effectively enhances adversarial robustness while preserving high clean accuracy across different model architectures and datasets, successfully addressing the accuracy-robustness trade-off challenge.

\noindent\textbf{Results on Tiny-ImageNet Dataset.} To further evaluate the scalability of our proposed method to more complex datasets, we conduct extensive experiments on the Tiny-ImageNet dataset. Table~\ref{tab_white_eval_Tiny-ImageNet} presents the white-box robustness results using both PreActResNet-18 (RN-18) and MobileNet-V2 (MN-V2) architectures, while Table~\ref{tab_black_eval_Tiny-ImageNet} displays the black-box robustness results against various attacks.

On the Tiny-ImageNet dataset, CIARD achieves state-of-the-art performance with 57.42\% clean accuracy and 28.26\% robust accuracy under FGSM attacks for PreActResNet-18, surpassing B-MTARD by 0.61\% and 0.14\% respectively. Similar improvements are observed with MobileNet-V2, where CIARD achieves 53.05\% clean accuracy and 25.45\% robust accuracy, outperforming B-MTARD (52.98\%, 25.60\%).

For black-box attacks, CIARD demonstrates even more pronounced advantages. Under PGDtrades attack, CIARD achieves 36.80\% robust accuracy and 47.11\% weighted robust accuracy with PreActResNet-18, exceeding B-MTARD's performance (36.65\%, 46.73\%). Against the more challenging Square Attack (SA), CIARD maintains superior performance with 44.48\% robust accuracy and 50.95\% weighted robust accuracy, compared to B-MTARD's 44.46\% and 50.64\% respectively.

These comprehensive results on the more complex Tiny-ImageNet dataset further confirm the effectiveness and scalability of our approach across different network architectures and attack types, demonstrating CIARD's ability to maintain the balance between clean accuracy and adversarial robustness even on more challenging visual recognition tasks.

\begin{table*}[htbp]
\centering
\caption{White-box Adversarial Robustness of ResNet-18 on CIFAR-10 and CIFAR-100 Datasets. The best results are \textbf{bolded}, and the second best results are \uline{underlined}.}
\label{tab_white_box_cifar10_cifar100_ResNet-18_complete}
\newcommand{\bestval}[1]{\textbf{#1}}
\newcommand{\secondval}[1]{\uline{#1}}
\small
\setlength{\tabcolsep}{4pt}
\begin{tabular}{c|c|c|l|ccc|ccc}
\toprule
\multirow{2}{*}{Student Model} & \multirow{2}{*}{Attack} & \multirow{2}{*}{Type} & \multirow{2}{*}{Defense} & 
\multicolumn{3}{c|}{CIFAR-10} & \multicolumn{3}{c}{CIFAR-100} \\
\cmidrule(lr){5-7} \cmidrule(lr){8-10}
& & & & Clean & Robust & W-Robust & Clean & Robust & W-Robust \\
\midrule
\multirow{10}{*}{ResNet-18} & \multirow{10}{*}{FGSM~\cite{goodfellow2014explaining}} & \multirow{7}{*}{Single-Teacher}
& SAT~\cite{madry2017towards}               &84.20  &55.59  &69.90 & 56.16 & 25.88 & 41.02 \\
& & & TRADES~\cite{zhang2019theoretically}  &83.00  &58.35  &70.68 & 57.75 & 31.36 & 44.56 \\
& & & ARD~\cite{goldblum2020adversarially}  &84.11  &58.40  &71.26 & 60.11 & 33.61 & 46.86 \\
& & & RSLAD~\cite{zi2021revisiting}         &83.99  &60.41  &72.20 & 58.25 & \secondval{34.73} & 46.49 \\
& & & SCORE~\cite{pang2022robustness}       &84.43  &59.84  &72.14 & 56.40 & 32.94 & 44.67 \\
& & & Fair-ARD~\cite{yue2024revisiting}     &83.41  &58.91  &71.16 & 57.81 & 34.39 & 46.10 \\
& & & ABSLD~\cite{zhao2023improving}        &83.21  &60.22  &71.72 & 56.77 & \bestval{34.94} & 45.86 \\
\cmidrule{3-10}
& & \multirow{3}{*}{Dual-Teacher} & MTARD~\cite{zhao2022enhanced}  & 87.36 & 61.20 & 74.28 & 64.30 & 31.49 & 47.90 \\
& & & B-MTARD~\cite{zhao2024mitigating}    & \secondval{88.20} & \secondval{61.42} & \secondval{74.81} & \secondval{65.08} & 34.21 & \secondval{49.65} \\
& & & CIARD      & \bestval{88.87} & \bestval{61.88} & \bestval{75.38} & \bestval{65.73} & 34.47 & \bestval{50.10} \\
\midrule
\multirow{10}{*}{ResNet-18} & \multirow{10}{*}{PGD$_\text{sat}$~\cite{madry2017towards}} & \multirow{7}{*}{Single-Teacher}
& SAT~\cite{madry2017towards}               &84.20  &45.85  &65.08 & 56.16 & 21.18 & 38.67 \\
& & & TRADES~\cite{zhang2019theoretically}  &83.00  &52.35  &67.68 & 57.75 &28.05 &42.90 \\
& & & ARD~\cite{goldblum2020adversarially}  &84.11  &50.93  &67.52 & 60.11 & 29.40 &44.76 \\
& & & RSLAD~\cite{zi2021revisiting}         &83.99  &\secondval{53.94}  &68.97 & 58.25 &\secondval{31.19} &44.72 \\
& & & SCORE~\cite{pang2022robustness}       &84.43  &53.72  &69.08 & 56.40 &30.27 &43.34 \\
& & & Fair-ARD~\cite{yue2024revisiting}     &83.41  &52.00  &67.71 & 57.81 &30.64 &44.23 \\
& & & ABSLD~\cite{zhao2023improving}        &83.21  &\bestval{54.63}  &68.92 & 56.77 &\bestval{32.41} &44.59\\
\cmidrule{3-10}
& & \multirow{3}{*}{Dual-Teacher} & MTARD~\cite{zhao2022enhanced}  & 87.36 & 50.83 & 69.05 & 64.30 &24.95 &44.63 \\
& & & B-MTARD~\cite{zhao2024mitigating}    & \secondval{88.20} & 51.68 & \secondval{69.94} & \secondval{65.08} &28.50 &\secondval{46.79} \\
& & & CIARD      & \bestval{88.87} & 51.70 & \bestval{70.29} & \bestval{65.73} & 28.05 & \bestval{46.89} \\
\midrule
\multirow{10}{*}{ResNet-18} & \multirow{10}{*}{PGD$_\text{trades}$~\cite{zhang2019theoretically}} & \multirow{7}{*}{Single-Teacher}
& SAT~\cite{madry2017towards}               &84.20  &48.12  &66.16 & 56.16 &22.02 &39.09 \\
& & & TRADES~\cite{zhang2019theoretically}  &83.00  &53.83  &68.42 & 57.75 &28.88 &43.32 \\
& & & ARD~\cite{goldblum2020adversarially}  &84.11  &52.96  &68.54 & 60.11 &30.51 &45.31 \\
& & & RSLAD~\cite{zi2021revisiting}         &83.99  &\secondval{55.73}  &69.86 & 58.25 &\secondval{32.05} &45.15 \\
& & & SCORE~\cite{pang2022robustness}       &84.43  &55.21  &69.82 & 56.40 &30.56 &43.48 \\
& & & Fair-ARD~\cite{yue2024revisiting}     &83.41  &53.77  &68.59 & 57.81  &31.50 &44.66 \\
& & & ABSLD~\cite{zhao2023improving}        &83.21  &\bestval{56.10}  &69.66 & 56.77 &\bestval{32.99} &44.88\\
\cmidrule{3-10}
& & \multirow{3}{*}{Dual-Teacher} & MTARD~\cite{zhao2022enhanced}  & 87.36 & 53.60 & 70.48 & 64.30 &26.75 &45.53 \\
& & & B-MTARD~\cite{zhao2024mitigating}    & \secondval{88.20} & 54.40 & \secondval{71.30} & 65.08 &29.94 &\secondval{47.51} \\
& & & CIARD      & \bestval{88.87} & 54.46 & \bestval{71.67} & \bestval{65.73} & 29.45 & \bestval{47.59} \\
\midrule
\multirow{10}{*}{ResNet-18} & \multirow{10}{*}{CW$_{\infty}$~\cite{carlini2017towards}} & \multirow{7}{*}{Single-Teacher}
& SAT~\cite{madry2017towards}               &84.20  &45.97  &65.09 & 56.16 &20.90 &38.53 \\
& & & TRADES~\cite{zhang2019theoretically}  &83.00  &50.23  &66.62 & 57.75 &24.19 &40.97 \\
& & & ARD~\cite{goldblum2020adversarially}  &84.11  &50.15  &67.13 & 60.11 &27.56 &43.84 \\
& & & RSLAD~\cite{zi2021revisiting}         &83.99  &\bestval{52.67}  &68.33 & 58.25  &\bestval{28.21} &43.23\\
& & & SCORE~\cite{pang2022robustness}       &84.43  &50.46  &67.45 & 56.40 &26.30 &41.35 \\
& & & Fair-ARD~\cite{yue2024revisiting}     &83.41  &51.07  &67.24 & 57.81  &\secondval{27.84} &42.83 \\
& & & ABSLD~\cite{zhao2023improving}        &83.21  &\secondval{52.04}  &67.63 & 56.77 & 26.99 &41.88 \\
\cmidrule{3-10}
& & \multirow{3}{*}{Dual-Teacher} & MTARD~\cite{zhao2022enhanced}  & 87.36 & 48.57 & 67.97 & 64.30 &23.42 &43.86 \\
& & & B-MTARD~\cite{zhao2024mitigating}    & \secondval{88.20} & 49.88 & \secondval{69.04} & \secondval{65.08} &25.45 &\bestval{45.27} \\
& & & CIARD      & \bestval{88.87} & 50.61 & \bestval{69.74} & \bestval{65.73} & 24.43 & \secondval{45.08} \\
\bottomrule
\end{tabular}
\end{table*}

\begin{table*}[htbp]
\centering
\caption{White-box Adversarial Robustness of MobileNet-V2 on CIFAR-10 and CIFAR-100 Datasets. The best results are \textbf{bolded}, and the second best results are \uline{underlined}.}
\label{tab_eval_White_cifar10_cifar100_MobileNet-V2}
\newcommand{\bestval}[1]{\textbf{#1}}
\newcommand{\secondval}[1]{\uline{#1}}
\small
\setlength{\tabcolsep}{4pt}
\begin{tabular}{c|c|c|l|ccc|ccc}
\toprule
\multirow{2}{*}{Student Model} & \multirow{2}{*}{Attack} & \multirow{2}{*}{Type} & \multirow{2}{*}{Defense} & 
\multicolumn{3}{c|}{CIFAR-10} & \multicolumn{3}{c}{CIFAR-100} \\
\cmidrule(lr){5-7} \cmidrule(lr){8-10}
& & & & Clean & Robust & W-Robust & Clean & Robust & W-Robust \\
\midrule
\multirow{10}{*}{MobileNet-V2} & \multirow{10}{*}{FGSM~\cite{goodfellow2014explaining}} & \multirow{7}{*}{Single-Teacher}
& SAT~\cite{madry2017towards}               &83.87  &55.89  &69.88 &59.19 &30.88 &45.04 \\
& & & TRADES~\cite{zhang2019theoretically}  &77.95  &53.75  &65.85 &55.41 &30.28 &42.85 \\
& & & ARD~\cite{goldblum2020adversarially}  & 83.43 & 57.03 & 70.23 &60.45 &32.77 &46.61 \\
& & & RSLAD~\cite{zi2021revisiting}         & 83.20  & \textbf{59.47} & 71.34 &59.01 &33.88 &46.45 \\
& & & SCORE~\cite{pang2022robustness}       & 82.32  & 58.43 & 70.38 &49.38 &29.28 &39.33 \\
& & & Fair-ARD~\cite{yue2024revisiting}     & 82.65  & 56.37 & 69.51 &59.18 &\secondval{34.07} &46.63 \\
& & & ABSLD~\cite{zhao2023improving}        & 82.50  & 58.47 & 70.49 &56.67 &33.85 &45.26 \\
\cmidrule{3-10}
& & \multirow{3}{*}{Dual-Teacher} & MTARD~\cite{zhao2022enhanced}  & \uline{89.26}  & 57.84 & 73.55 &\bestval{67.01} &32.42 &49.72 \\
& & & B-MTARD~\cite{zhao2024mitigating}     & 89.09  & 58.79 & \uline{73.94} &66.13 &\bestval{34.36} &\bestval{50.25}\\
& & & CIARD      & \textbf{89.51}  & \uline{59.10}  & \textbf{74.31} & \secondval{66.72} & 33.56 & \secondval{50.14} \\
\midrule
\multirow{10}{*}{MobileNet-V2} & \multirow{10}{*}{PGD$_\text{sat}$~\cite{madry2017towards}} & \multirow{7}{*}{Single-Teacher}
& SAT~\cite{madry2017towards}               &83.87  &46.84  &65.36 &59.19 &25.64 &42.42 \\
& & & TRADES~\cite{zhang2019theoretically}  &77.95  &49.06  &63.51 &55.41 &23.33 &39.37 \\
& & & ARD~\cite{goldblum2020adversarially}  &83.43 &49.50 & 66.47 &60.45 &28.69 &44.57 \\
& & & RSLAD~\cite{zi2021revisiting}         &83.20 &\uline{53.25} &68.23 &59.01 &\secondval{30.19} &44.60 \\
& & & SCORE~\cite{pang2022robustness}       & 82.32 &\textbf{53.42} & 67.87 &49.38 &27.03 &38.21\\
& & & Fair-ARD~\cite{yue2024revisiting}     & 82.65 &50.50 &66.58 &59.18 &30.15 &44.67 \\
& & & ABSLD~\cite{zhao2023improving}        & 82.50 &52.98 &67.74 &56.67 &\bestval{31.28} &43.98 \\
\cmidrule{3-10}
& & \multirow{3}{*}{Dual-Teacher} & MTARD~\cite{zhao2022enhanced}  & \uline{89.26}  &44.16 &66.71 &\bestval{67.01} &25.14 &46.08 \\
& & & B-MTARD~\cite{zhao2024mitigating}     & 89.09  & 47.56 & \uline{68.33} &66.13 &28.47 &\bestval{47.30} \\
& & & CIARD      & \textbf{89.51}  & 47.67  & \textbf{68.59} & \secondval{66.72} & 27.02 & \secondval{46.87} \\
\midrule
\multirow{10}{*}{MobileNet-V2} & \multirow{10}{*}{PGD$_\text{trades}$~\cite{zhang2019theoretically}} & \multirow{7}{*}{Single-Teacher}
& SAT~\cite{madry2017towards}               &83.87  &49.14  &66.51 &59.19 &26.96 &43.08 \\
& & & TRADES~\cite{zhang2019theoretically}  &77.95  &50.27  &64.11 &55.41 &28.42 &41.92 \\
& & & ARD~\cite{goldblum2020adversarially}  & 83.43 &51.70 & 67.57 &60.45 &29.63 &45.04 \\
& & & RSLAD~\cite{zi2021revisiting}         & 83.20 &\textbf{54.76} & 68.98 &59.01 &31.19 &45.10 \\
& & & SCORE~\cite{pang2022robustness}       & 82.32 &54.46 & 68.39 &49.38 &27.53 &38.46 \\
& & & Fair-ARD~\cite{yue2024revisiting}     & 82.65 &52.12 &67.39 &59.18 &\secondval{31.26} &45.22 \\
& & & ABSLD~\cite{zhao2023improving}        & 82.50 &\uline{54.49} &68.50 &56.67 &\bestval{31.90} &44.29 \\
\cmidrule{3-10}
& & \multirow{3}{*}{Dual-Teacher} & MTARD~\cite{zhao2022enhanced}  & \uline{89.26}  &47.99 &68.63 &\bestval{67.01} &27.10 &47.06 \\
& & & B-MTARD~\cite{zhao2024mitigating}     & 89.09  & 50.44 & \uline{69.77} &66.13 &29.82 &\bestval{47.98} \\
& & & CIARD      & \textbf{89.51}  & 50.71  & \textbf{70.11} & \secondval{66.72} & 28.95 & 
\secondval{47.84}\\
\midrule
\multirow{10}{*}{MobileNet-V2} & \multirow{10}{*}{CW$_{\infty}$~\cite{carlini2017towards}} & \multirow{7}{*}{Single-Teacher}
& SAT~\cite{madry2017towards}               &83.87  &46.62  &65.25 &59.19 &25.01 &42.10 \\
& & & TRADES~\cite{zhang2019theoretically}  &77.95  &46.06  &62.01 &55.41 &\secondval{27.72} &41.57 \\
& & & ARD~\cite{goldblum2020adversarially}  & 83.43 &48.96 & 66.20 &60.45 &26.55 &43.50 \\
& & & RSLAD~\cite{zi2021revisiting}         & 83.20 &\textbf{51.78} & 67.49&59.01 &\bestval{27.98} &43.50 \\
& & & SCORE~\cite{pang2022robustness}       & 82.32 &49.18 & 65.75 &49.38 &23.29 &36.34 \\
& & & Fair-ARD~\cite{yue2024revisiting}     & 82.65 &\uline{51.07} &66.86 &59.18 &27.55 &43.37\\
& & & ABSLD~\cite{zhao2023improving}        & 82.50 &50.20 &66.35 &56.67 &26.40 &41.54 \\
\cmidrule{3-10}
& & \multirow{3}{*}{Dual-Teacher} & MTARD~\cite{zhao2022enhanced}  & \uline{89.26}  &43.42 &66.34 &\bestval{67.01} &24.14 &45.58 \\
& & & B-MTARD~\cite{zhao2024mitigating}     & 89.09  & 46.81 & \uline{67.95} &66.13 &26.50 &\bestval{46.32} \\
& & & CIARD      & \textbf{89.51}  & 46.88  & \textbf{68.20} & \secondval{66.72} & 25.54 & \secondval{46.13} \\
\bottomrule
\end{tabular}
\end{table*}

\begin{table}[htbp]\small
\centering
\begin{center}
\caption{The white-box robustness of the Tiny-ImageNet dataset is tested using PreActResNet-18 (RN-18) and MobileNet-V2 (MN-V2), respectively.}
\label{tab_white_eval_Tiny-ImageNet}
\newcommand{\bestval}[1]{\textbf{#1}}
\newcommand{\secondval}[1]{\uline{#1}}
\vspace{-0.1in}
\setlength{\tabcolsep}{0.05mm}{
\begin{tabular}{l|l|ccc|ccc}
\toprule
\multirow{2}{*}{Attack} & \multirow{2}{*}{Defense} & \multicolumn{3}{c|}{Tiny-ImageNet(RN-18)} & \multicolumn{3}{c}{Tiny-ImageNet(MN-V2)} \\
 &  & Clean & Robust & W-Robust & Clean & Robust & W-Robust \\
\midrule
\multirow{10}{*}{FGSM} 
& SAT        & 50.08 &25.35 &37.72                    &49.03 &23.38 &36.21 \\
& TRADES     & 48.45 &23.59 &36.02                    &43.81 &20.10 &31.96 \\
& ARD        &53.22 &27.97 &40.60                     &45.53 &22.88 &33.21 \\
& RSLAD      &48.78 &27.26 &38.02                     &45.69 &24.09 &34.89 \\
& SCORE      &10.05 &7.80 &8.93                       &28.27 &17.47 &22.87 \\
& Fair-ARD   &46.64 &25.81 &36.23                     &47.24 &25.31 &36.28 \\
& MTARD      &52.98 &26.41 &39.70                    &50.50 &23.94 &37.22 \\
& B-MTARD    &\secondval{56.81} &\secondval{28.12} &\secondval{42.47}                    &\secondval{52.98} &\bestval{25.60} &\bestval{39.29}  \\
& CIARD      & \bestval{57.42}  & \bestval{28.26}  &\bestval{42.84}                & \bestval{53.05}  & \secondval{25.45}  & \secondval{39.25}  \\
\cmidrule{1-8}
\multirow{10}{*}{PGD$_\text{sat}$} 
& SAT        &50.08  &22.24 &36.16                     &49.03 &20.31 &34.67 \\
& TRADES     &48.45 &21.59 &35.02                     &43.81 &18.16 &30.99 \\
& ARD        &53.22 &\bestval{24.92} &39.07                     &45.53 &20.43 &32.98 \\
& RSLAD      &48.78 &25.00 &36.89                     &45.69 &\secondval{22.30} &34.00 \\
& SCORE      &10.05 &7.65 &8.85                       &28.27 &16.48 &22.38 \\
& Fair-ARD   &46.64 &23.91 &35.28                     &47.24 &\bestval{23.37} &35.31 \\
& MTARD      &52.98 &22.55 &37.77                     &50.50 &20.45 &35.48  \\
& B-MTARD    &\secondval{56.81} &23.93 &\secondval{40.37}                     &\secondval{52.98} &21.58 &\bestval{37.28}  \\
& CIARD     & \bestval{57.42}  & \secondval{23.95}  & \bestval{40.69}               & \bestval{53.05}  & 21.38  &\secondval{37.22}   \\
\cmidrule{1-8}
\multirow{10}{*}{PGD$_\text{trades}$} 
& SAT        &50.08  &23.05 &36.57                     &49.03 &21.15 &35.09 \\
& TRADES     &48.45 &22.09 &35.27                     &43.81 &18.36 &31.09 \\
& ARD        &53.22  &\bestval{25.71} &39.47                     &45.53 &21.00 &33.27 \\
& RSLAD      &48.78 &\secondval{25.45} &37.12                     &45.69 &\secondval{22.74} &34.22 \\
& SCORE      &10.05  &7.67 &8.86                    &28.27 &16.69 &22.48\\
& Fair-ARD   &46.64 &24.29 &35.47                     &47.24 &\bestval{23.77} &35.51 \\
& MTARD      &52.98 &23.41 &38.20                     &50.50 &21.20 &35.85 \\
& B-MTARD    &\secondval{56.81} &24.94 &\secondval{40.88}                     &\secondval{52.98} &22.58 &\bestval{37.78}  \\
& CIARD      & \bestval{57.42}  & 24.89  & \bestval{41.16}                 & \bestval{53.05}  & 22.42  & \secondval{37.74} \\
\cmidrule{1-8}
\multirow{10}{*}{CW$_{\infty}$} 
& SAT        &50.08 &20.48 &35.28                     &49.03 &\secondval{18.69} &33.86 \\
& TRADES     &48.45 &17.33 &32.89                    &43.81 &13.47 &28.66 \\
& ARD        &53.22 &\bestval{21.41} &37.32                     &45.53 &16.81 &31.17 \\
& RSLAD      &48.78 &\secondval{20.87} &34.83                    &45.69 &18.63 &32.16 \\
& SCORE      &10.05 & 6.19 &8.13                     &28.27 &13.25 &20.76 \\
& Fair-ARD   &46.64 &19.59 &33.12                     &47.24 &\bestval{20.04} &33.64 \\
& MTARD      &52.98 &19.36 &36.17                     &50.50  &17.45 &33.98  \\
& B-MTARD    &\secondval{56.81} &19.69 &\secondval{38.25}                     &\secondval{52.98} &18.08 &\secondval{35.53}  \\
& CIARD      & \bestval{57.42}  & 19.56  & \bestval{38.49}                 & \bestval{53.05}  & 18.20  & \bestval{35.63}\\
\cmidrule{1-8}
\end{tabular}
}
\vspace{-0.075in}
\end{center}
\end{table}

\begin{table}[htbp]\small
\centering
\begin{center}
\caption{Black-box Adversarial Robustness of MobileNet-V2 on CIFAR-10 and CIFAR-100 Datasets.}
\label{tab_black_eval_Tiny-ImageNet}
\newcommand{\bestval}[1]{\textbf{#1}}
\newcommand{\secondval}[1]{\uline{#1}}
\vspace{-0.075in}
\setlength{\tabcolsep}{0.1mm}{
\begin{tabular}{l|l|ccc|ccc}
\toprule
\multirow{2}{*}{Attack} & \multirow{2}{*}{Defense} & \multicolumn{3}{c|}{Tiny-ImageNet(RN-18)} & \multicolumn{3}{c}{Tiny-ImageNet(MN-V2)} \\
 &  & Clean & Robust & W-Robust & Clean & Robust & W-Robust \\
\midrule
\multirow{10}{*}{PGD$_\text{trades}$} 
& SAT        &50.08  &33.40 &41.74                     &49.03 &33.47 &41.25 \\
& TRADES     &48.45 &31.01 &39.73                     &43.81 &28.35 &36.08 \\
& ARD        &53.22 &34.74 &43.98                     &45.53 &30.73 &38.13 \\
& RSLAD      &48.78 &32.85 &40.82                     &45.69 &31.20 &38.45 \\
& SCORE      &10.05 &8.74 &9.40                       &28.27 &21.82 &25.05 \\
& Fair-ARD   &46.64 &31.58 &39.11                     &47.24 &31.80 &39.52 \\
& ABSLD      &47.21 &31.84 &39.53                     &48.08 &32.89 &40.49 \\
& MTARD      &52.98  &34.48 &43.73                     &50.50 &32.75 &41.63  \\
& B-MTARD    &\secondval{56.81} &\secondval{36.65} &\secondval{46.73}                     &\secondval{52.98} &\secondval{34.25} &\secondval{43.62}  \\
& CIARD     & \bestval{57.42}  & \bestval{36.80}  & \bestval{47.11}                 & 
\bestval{53.05}  & \bestval{34.50}  & \bestval{43.78}  \\
\cmidrule{1-8}
\multirow{10}{*}{CW$_{\infty}$} 
& SAT        &50.08  &33.20 &41.63                     &49.03 &33.13 &41.08 \\
& TRADES     &48.45 &30.72 &39.59                     &43.81 &28.64 &36.23 \\
& ARD        &53.22 &33.32 &43.27                     &45.53 &30.23 &37.88 \\
& RSLAD      &48.78 &32.09 &40.44                     &45.69 &31.10 &38.40 \\
& SCORE      &10.05 &8.82 &9.44                       &28.27 &22.19 &25.23 \\
& Fair-ARD   &46.64 &31.38 &39.01                     &47.24 &31.40 &39.32 \\
& ABSLD      &47.21 &31.66 &39.44                     &48.08 &32.43 &40.26 \\
& MTARD      &52.98 &\bestval{33.80} &43.39                     &50.50 &32.05 &41.28  \\
& B-MTARD    &\secondval{56.81} &33.40 &\secondval{45.11}                     &\secondval{52.98} &\bestval{33.50} &\secondval{43.24}  \\
& CIARD     & \bestval{57.42} & \secondval{33.69}  & \bestval{45.56}                 & \bestval{53.05}  & \secondval{33.45}  & \bestval{43.25}  \\
\cmidrule{1-8}
\multirow{10}{*}{SA~\cite{andriushchenko2020square}} 
& SAT        &50.08 &38.72 &44.40                     &49.03 &37.95 &43.49 \\
& TRADES     &48.45 &36.58 &42.52                     &43.81 &32.39 &38.10 \\
& ARD        &53.22 &42.58 &47.90                     &45.53 &34.60 &40.07 \\
& RSLAD      &48.78 &37.64 &43.21                     &45.69 &35.18 &40.44 \\
& SCORE      &10.05 & 8.67 &9.36                       &28.27 &22.16 &25.22 \\
& Fair-ARD   &46.64 &35.81 &41.23                     &47.24 &36.53 &41.89 \\
& ABSLD      &47.21 &36.77 &41.99                     &48.08 &38.18 &43.13 \\
& MTARD      &52.98 &41.70 &47.34                     &50.50 &38.88 &44.69  \\
& B-MTARD    &\secondval{56.81} &\secondval{44.46} &\secondval{50.64}                     &\secondval{52.98} &\bestval{40.62} &\bestval{46.80}  \\
& CIARD     & \bestval{57.42} & \bestval{44.48}  & \bestval{50.95}                 & \bestval{53.05}  & \secondval{39.96}  & \secondval{46.51}  \\
\cmidrule{1-8}
\end{tabular}
}
\vspace{-0.075in}
\end{center}
\end{table}

\noindent \textbf{Robustness Evaluation against AutoAttack.} To provide a more comprehensive and rigorous evaluation of adversarial robustness, we tested our proposed method against \textbf{AutoAttack}~\cite{croce2020reliable}, which is widely considered a state-of-the-art and parameter-free ensemble of attacks. This evaluation provides a reliable estimate of a model's worst-case robustness under strong, adaptive threat models, thereby addressing the need for evaluation beyond standard PGD-based attacks. The evaluation was conducted on the CIFAR-10 dataset using both ResNet-18 and MobileNetV2 as student architectures. The results, including clean accuracy, robust accuracy, and the holistic Weighted Robustness (W-R) metric, are presented in Table~\ref{tab:autoattack_summary}.

\begin{table*}[htbp]\small
\centering
\caption{Performance evaluation against AutoAttack on CIFAR-10. We report clean accuracy (Clean), robust accuracy (Robust), and Weighted Robustness (W-R). The \textbf{best} results are in bold, and the \underline{second-best} are underlined. Our CIARD method achieves the highest W-R on both architectures.}
\label{tab:autoattack_summary}

\begin{tabular}{l|ccc|ccc}
\toprule
\multirow{2}{*}{\textbf{Defense Method}} & \multicolumn{3}{c|}{\textbf{ResNet-18}} & \multicolumn{3}{c}{\textbf{MobileNetV2}} \\
\cmidrule(lr){2-4} \cmidrule(lr){5-7}
& Clean (\%) & Robust (\%) & W-R (\%) & Clean (\%) & Robust (\%) & W-R (\%) \\
\midrule
RSLAD      & 83.99          & \textbf{50.98} & 67.49          & 83.20          & \textbf{50.23} & 66.72 \\
Fair-ARD   & 83.41          & 49.21          & 66.31          & 82.65          & 47.68          & 65.17 \\
ABSLD      & 83.21          & \underline{50.60} & 66.91          & 82.50          & \underline{48.65} & 65.58 \\
MTARD      & 87.36          & 46.18          & 66.77          & \textbf{89.26} & 41.02          & 65.14 \\
B-MTARD    & \underline{88.20} & 47.44          & \underline{67.82} & \underline{89.09} & 44.58          & \underline{66.84} \\
\midrule
\textbf{CIARD (Ours)} & \textbf{88.87} & 48.88          & \textbf{68.88} & 88.90          & 46.31          & \textbf{67.61} \\
\bottomrule
\end{tabular}%
\end{table*}

The results clearly demonstrate the superiority of our proposed CIARD framework. It achieves the highest Weighted Robustness (W-R) on both architectures, reaching \textbf{68.88\% on ResNet-18} and \textbf{67.61\% on MobileNetV2}. This state-of-the-art overall performance highlights its exceptional ability to balance high accuracy on clean samples with strong defense against adversarial attacks. Compared to the strong B-MTARD baseline, CIARD yields a significant W-R improvement of \textbf{+1.06\%} on ResNet-18 and \textbf{+0.77\%} on MobileNetV2. It is worth noting that while a specialized method like RSLAD achieves the highest raw robust accuracy, it does so at the cost of a considerable drop in clean accuracy. In contrast, CIARD maintains a high clean accuracy (e.g., 88.87\% on ResNet-18) while delivering competitive robustness. This confirms that our proposed contrastive push loss and iterative teacher training are highly effective at mitigating the common accuracy-robustness trade-off. In summary, the strong performance under the demanding AutoAttack benchmark further validates the effectiveness of CIARD in producing lightweight models that are both robust and accurate, making them more reliable for real-world deployment.

\noindent \textbf{Ablation Study on the Push Loss Weight \texorpdfstring{$\lambda$}{lambda}.} Our proposed CIARD framework introduces a key hyperparameter, $\lambda$, which controls the weight of the contrastive push loss term responsible for robust specialization. To analyze the sensitivity of our method to this parameter, we conducted a dedicated ablation study. In our main experiments, we set $\lambda=1.0$ by default. For this analysis, to isolate the effect of $\lambda$, we kept the other primary loss weights fixed at $\alpha=1$ and $\beta=1$. The study was performed using the ResNet-18 student model on the CIFAR-10 dataset.

The results, presented in Table~\ref{tab:lambda_ablation}, show the model's performance under various white-box attacks and the comprehensive AutoAttack suite as $\lambda$ is varied.

\begin{table*}[htbp]\small
\centering
\caption{Ablation study on the push loss weight $\lambda$. Experiments were run with ResNet-18 on CIFAR-10. We report clean accuracy and robust accuracy under multiple attack scenarios. The setting used in our main experiments ($\lambda=1.0$) is highlighted.}
\label{tab:lambda_ablation}
\begin{tabular}{c|c|cccc|c}
\toprule
\textbf{Weight} $\boldsymbol{\lambda}$ & \textbf{Clean (\%)} & \textbf{PGD-T (\%)} & \textbf{PGD-S (\%)} & \textbf{FGSM (\%)} & \textbf{CW}$\boldsymbol{_\infty}$ \textbf{(\%)} & \textbf{AutoAttack (\%)} \\
\midrule
0.8 & 89.06 & 53.85 & 51.21 & 61.12 & 50.31 & 48.48 \\
\textbf{1.0} & \textbf{88.87} & \textbf{54.46} & \textbf{51.70} & \textbf{61.88} & \textbf{50.61} & \textbf{48.88} \\
1.2 & 88.60 & 54.55 & 51.98 & 61.81 & 50.92 & 49.25 \\
\bottomrule
\end{tabular}%
\end{table*}

The findings reveal a clear and expected trade-off. As $\lambda$ increases, a greater emphasis is placed on pushing the student's predictions away from the clean teacher's vulnerabilities, which generally leads to improved adversarial robustness. This is evidenced by the rising robust accuracy against most attacks, particularly the strong AutoAttack benchmark (from 48.48\% to 49.25\%). This gain in robustness is accompanied by a slight and graceful degradation in clean accuracy (from 89.06\% to 88.60\%). Our chosen default value of $\lambda=1.0$ strikes an effective balance, achieving strong performance across both clean and adversarial examples. The relative stability of the results across the tested range also indicates that our method is not overly sensitive to this hyperparameter.
\section{Future Research Directions}
\label{Future_Research_Directions}

\noindent\textbf{Scaling of multi-modal data.} We can extend CIARD to multi-modal data, such as images, text and audio, to enhance the robustness of the model across different data modalities. The fusion and processing of multi-modal data is crucial for improving model robustness. Future research could focus on designing a unified framework that enables CIARD to process multiple data types and utilise the complementary information between these modalities to improve robustness and accuracy. This approach will significantly expand the applicability and effectiveness of CIARD in a variety of realistic scenarios.

\noindent\textbf{Cross-domain applications.} We can try to apply CIARD to various domains such as healthcare, finance, and autonomous driving to verify its versatility and usefulness. Different domains have unique data characteristics and application requirements, and future research could explore how CIARD can be adapted and optimised for these specific scenarios. By doing so, the effectiveness and value of CIARD can be validated in real-world applications, thus ensuring its wider applicability and impact in different domains.

\noindent\textbf{Integration with mainstream architectures.} In this paper, we focus on DNN models because they have established benchmarks and are widely used in current applications. However, we also recognize the importance of evaluating our approach on newer architectures. In our current work, we not only focus on the latest models like ViT and Swin-Transformer, but also combine the methods of this paper with LLMs to carry out NLP tasks as a way to improve the robustness of the target model.

Through these research directions, the performance and application scope of CIARD can be further improved, laying the foundation for building more robust and efficient edge intelligence systems.

{
    \small
    \normalem
    \bibliographystyle{ieeenat_fullname}
    \bibliography{main}
}

%% file: sec/0_abstract.tex
\begin{abstract}
Adversarial robustness distillation (ARD) aims to transfer both performance and robustness from teacher model to lightweight student model, enabling resilient performance on resource-constrained scenarios.
Though existing ARD approaches enhance student model's robustness, the inevitable by-product leads to the degraded performance on clean examples. We summarize the causes of this problem inherent in existing methods with dual-teacher framework as:  
\textcircled{1} The divergent optimization objectives of dual-teacher models, i.e., the clean and robust teachers, impede effective knowledge transfer to the student model, and \textcircled{2} The iteratively generated adversarial examples during training lead to performance deterioration of the robust teacher model. To address these challenges, we propose a novel Cyclic Iterative ARD (CIARD) method with two key innovations: \textcircled{1} A multi-teacher framework with contrastive push-loss alignment to resolve conflicts in dual-teacher optimization objectives, and \textcircled{2} Continuous adversarial retraining to maintain dynamic teacher robustness against performance degradation from the varying adversarial examples. Extensive experiments on CIFAR-10, CIFAR-100, and Tiny-ImageNet demonstrate that CIARD achieves remarkable performance with an average \textbf{3.53\%} improvement in adversarial defense rates across various attack scenarios and a \textbf{5.87\%} increase in clean sample accuracy, establishing a new benchmark for balancing model robustness and generalization. Our code is available at \href{https://github.com/eminentgu/CIARD}{\textcolor{red}{https://github.com/eminentgu/CIARD}}.
\end{abstract}

%% file: sec/1_intro.tex
\section{Introduction}
\label{sec:intro}


\blfootnote{$^{*}$Corresponding author.}
In the era of edge computing and real-time applications, the deployment of efficient and robust models on devices with limited resources presents a significant challenge. Knowledge distillation (KD)~\cite{hinton2015distilling,yang2023knowledge,lu2024uniads,zhou2021bert} has become a popular method for compressing large teacher models into smaller, more efficient student models with minimal accuracy loss. However, adversarial attacks~\cite{ma2021understanding,madry2017towards,li2022review,szegedy2013intriguing,papernot2016limitations,bai2023query} pose a significant threat to the deployment of these student models, particularly in edge environments. Therefore, enhancing the robustness of lightweight student models against adversarial attacks is crucial for their application in real-world scenarios such as autonomous driving, image classification, and speech recognition~\cite{eykholt2018robust,he2016deep,wang2017residual,lyu2024unibind}.


\begin{figure}[t]
    \centering
    \includegraphics[width=1\linewidth]{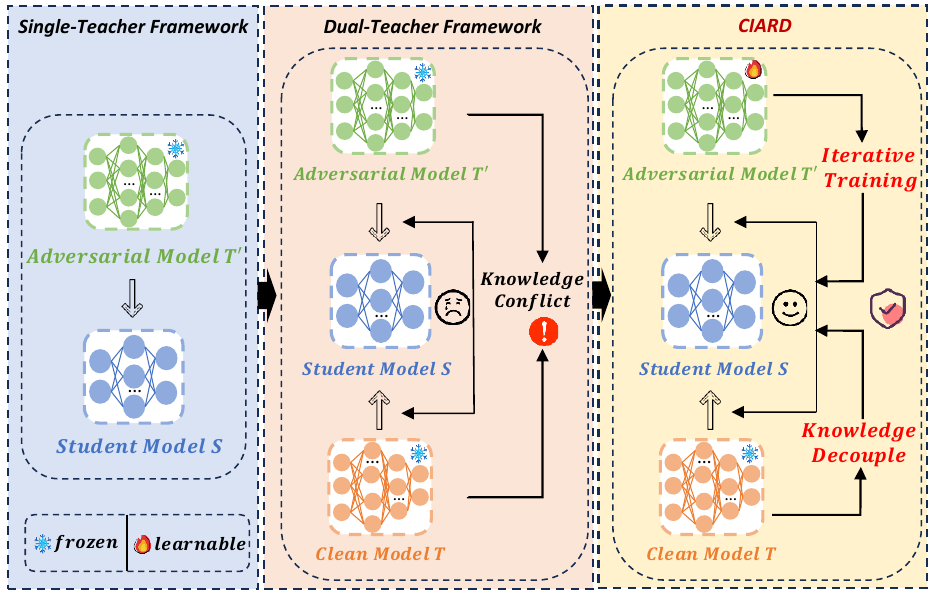}
    \vspace{-16pt}
    \caption{Comparison of single-teacher, dual-teacher and CIARD (ours) distillation frameworks.}
    \vspace{-8pt}
    \label{fig:illustration}
\end{figure}


To address these security concerns, adversarial training (AT)~\cite{ma2018characterizing,jia2022adversarial,Hsiung_2023_CVPR,10478545,10471619} has emerged as a promising defense approach. AT enhances model robustness~\cite{athalye2018obfuscated,croce2020reliable,wang2024dp,xu2024enhancing} by incorporating both clean and adversarial examples during the training process. However, when applied to knowledge distillation, AT presents two significant challenges including: \textcircled{1} Increasing computational burden for generating adversarial attacks contradicts the model compression efficiency, and \textcircled{2} The robustness-accuracy trade-off problem that models typically sacrifice clean performance for better robustness.

Recently, adversarial robustness distillation (ARD)~\cite{zi2021revisiting,huang2023boosting,yue2024revisiting,zhu2023improving,goldblum2020adversarially,zhang2019theoretically} has emerged as a promising solution. In the context of ARD, the dual-teacher architecture~\cite{zhao2022enhanced} offers powerful solution to the two aforementioned challenges 
by simultaneously improving both clean sample accuracy and adversarial robustness when guiding the student models. 
However, two critical challenges remain in the current dual-teacher ARD frameworks including:
\textcircled{1} \textbf{\textit{Conflicting Optimization Objectives}:} In conventional dual-teacher distillation frameworks, the two teachers serve distinct purposes where one teacher focuses on clean sample accuracy while the other emphasizes adversarial robustness. This dichotomy often leads to suboptimal knowledge transfer as the student model struggles to reconcile these competing objectives. As shown in Table~\ref{Table:simple_comparison}, current methods, such as B-MTARD \cite{zhao2024mitigating}, improve robustness ($\uparrow 0.95\% $) at the cost of a decrease in clean accuracy ($\downarrow 0.17\%$). This cost between clean accuracy and robust accuracy clearly validates the issue, and \textcircled{2} \textbf{\textit{Degradation of Adversarial Teacher Performance}:} Through empirical observations (Fig.~\ref{fig:Robust_Clean}), we find that as the student model evolves during training, the generated adversarial examples increasingly compromise the performance of the robust teacher model. This degradation significantly impacts the quality of knowledge transfer and the overall robustness of the student model.

\begin{figure*}[htbp]
    \centering
    \includegraphics[width=1\linewidth]{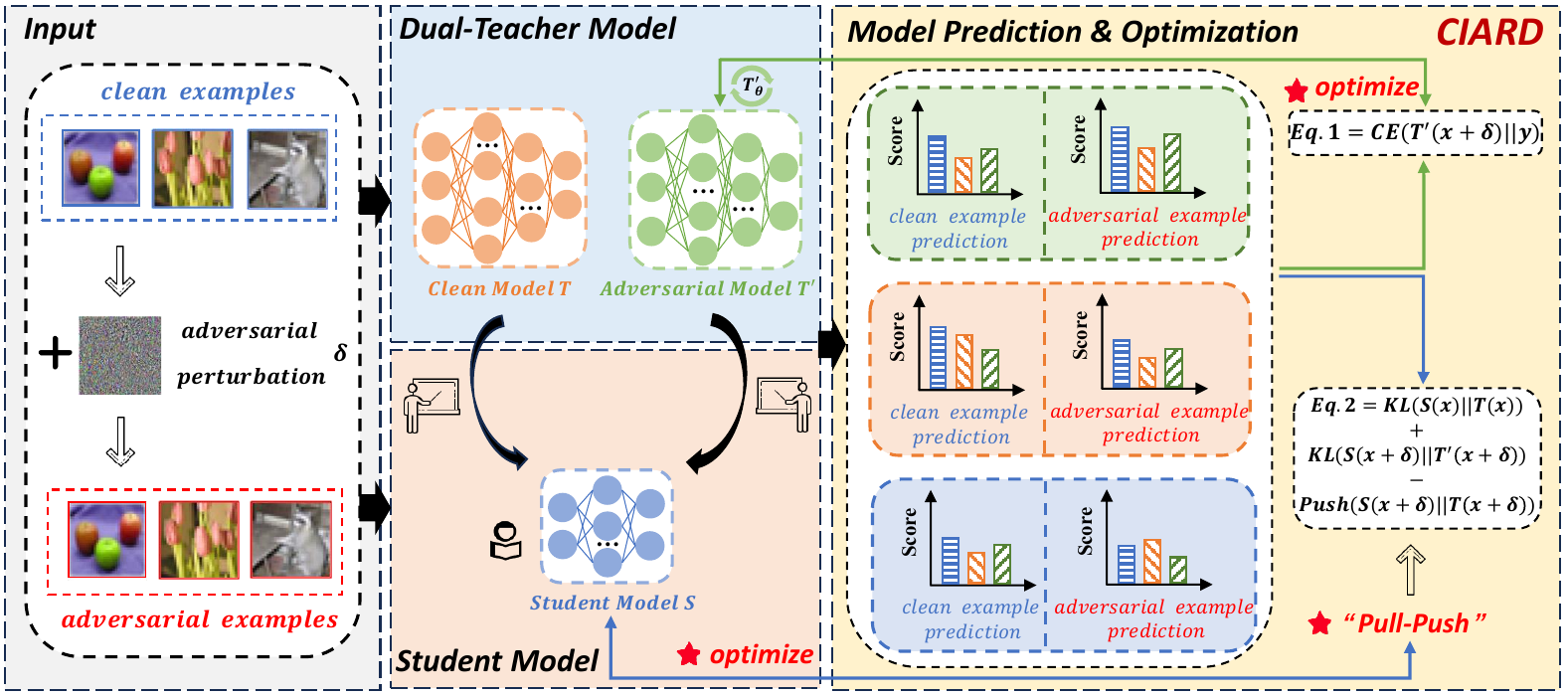}
    \vspace{-2em}
    \caption{The framework of Cyclic Iterative Adversarial Robustness Distillation (CIARD).  Our dual-teacher distillation framework features a continuously updated robust teacher model and a push loss mechanism to guide knowledge transfer, enabling effective balance between adversarial robustness and accuracy in the student model.}
    \label{fig:framework}
\end{figure*}

To address these limitations of prior dual-teacher ARDmethods~\cite{zhao2024mitigating,zhao2022enhanced}, we propose a novel cyclic iterative dual-teacher distillation framework, namely Cyclic Iterative Adversarial Robustness Distillation (CIARD), as shown in Figure~\ref{fig:illustration}.
The CIARD framework resolves the distinct training optimization objectives issue by introducing contrastive push loss alignment. 
The contrastive push loss 
effectively decouples clean knowledge from robust knowledge and ensure that student model specializes in learning robust features without interference from clean teacher. Practically, the push loss works by deliberately creating a divergence between the student and clean teacher. By pushing away from the clean teacher's incorrect predictions, the student effectively absorbs the robust teacher's specialized knowledge.

After achieving the goal of precisely guide the student's learning trajectory, we further incorporate an Iterative Teacher Training (ITT) strategy to
avoid the performance degradation of the teacher model caused by the iteratively generated adversarial examples during training. 
In detail, ITT first freezes both teacher models' parameters at the early training stage to let student have a basic knowledge. Afterwards, ITT iteratively updates the robustness teacher with our proposed continuous adversarial retraining to maintain dynamic teacher robustness against performance degradation from the varying adversarial examples.

Extensive experiments across multiple datasets demonstrate that CIARD significantly outperforms existing ARD methods, achieving substantial improvements in both adversarial robustness and clean sample accuracy.      
Overall, our contributions are summarized as follows:
\textbf{(I)} 
\textbf{(II)} We introduce a novel optimization strategy, namely Contrastive Push Loss, that resolves conflicting objectives among dual-teacher models, facilitating enhanced knowledge transfer while dynamically adjusting the training process to achieve an optimal balance between robustness and accuracy.
\textbf{(III)} We introduce the Iterative Teacher Training strategy to dynamically enhance the knowledge supplementation of the adversarial teacher based on the evaluation of the student model. This approach equips the robust teacher  a strong capability in both robust and clean knowledge, effectively boosting the student’s performance.
\textbf{(IV)} Extensive experiments across multiple datasets demonstrate that CIARD significantly outperforms existing ARD methods, achieving substantial improvements in both adversarial robustness and clean sample accuracy.

\begin{table}[t!]
\centering
\caption{Performance comparison of MobileNet-V2 under FGSM in two knowledge distillation frameworks on CIFAR-10. Best performance metrics are highlighted in \textbf{bold}.}
\vspace{-6pt}
\setlength{\tabcolsep}{4pt}
\label{Table:simple_comparison}
\resizebox{\linewidth}{!}{
\begin{tabular}{c|l|ll}
\toprule[1.5pt]
\multirow{2}{*}{Type} & \multirow{2}{*}{Defense} & \multicolumn{2}{c}{FGSM~\cite{goodfellow2014explaining}} \\
\cmidrule{3-4}
& & Clean (\%) & Robust (\%) \\ 
\midrule
\multirow{2}{*}{Single-Teacher} & ARD~\cite{goldblum2020adversarially}  & 83.43  & 57.03 \\
& ABSLD~\cite{zhao2023improving}    & 82.50  $\downarrow$ & 58.47 $\uparrow$ \\
\midrule
\multirow{3}{*}{Dual-Teacher} & MTARD~\cite{zhao2022enhanced} & 89.26 &  57.84 \\
& B-MTARD~\cite{zhao2024mitigating} & 89.09 $\downarrow$ & 58.79 $\uparrow$  \\
& \textbf{CIARD (Ours)} & \textbf{89.51} $\uparrow$ & \textbf{59.10} $\uparrow$\\
\bottomrule[1.5pt]
\end{tabular}}
\vspace{-12pt}
\end{table}

%% file: sec/2_preliminary_background.tex
\section{Preliminary \& Background}
\label{sec:preliminary_background}
\paragraph{Knowledge Distillation (KD)} 
KD is a widely used method for compressing deep neural networks, aiming to transfer knowledge from a large teacher model to a smaller student model~\cite{zheng2024learning,zhu2023good}. Given a teacher model $T(\cdot)$ and an in-distribution example $x \sim D$ (from the same distribution as the teacher’s training data), the traditional KD optimization objective is to obtain student model parameters $\theta$ that minimize the Kullback-Leibler (KL) divergence loss $\mathcal{L}_{KL}(\cdot)$ between the softmax logits of the teacher and student outputs. The loss function is defined as follows:
\begin{small}
\begin{align}
\label{TKD}
\arg\min_{\theta_{S}}\underbrace{(\alpha \text{CE}(S({x}),y))}_{\text{Hard Label Loss}} + \underbrace{(\beta \tau^{2}\text{KL}\left(S^{\tau}({x}),T^{\tau}({x})\right))}_{\text{Soft Label Loss}}.
\end{align}
\end{small}

As seen from the distillation Eq.~\ref{TKD}, the loss function typically consists of two components: one is the cross-entropy loss between the student model and the true label, and the other is the KL divergence between the student model and the teacher model's soft labels. However, traditional KD focuses solely on the student model's accuracy on clean examples, without considering its robustness.

\paragraph{Adversarial Training (AT)}
\label{sec:AT}
AT remains one of the most effective methods for defending against adversarial attacks. Its core principle involves training models with both adversarial and clean examples, enabling defense capabilities through exposure to hostile inputs~\cite{liao2025adversarial,xie2023adversarial}.
The key to adversarial training lies in generating perturbations through inner maximization. 
Extensive research has focused on enhancing the robustness of deep neural networks through AT, by proposing various strategies. For example, Zi et al.~\cite{zi2021revisiting} employ robust soft labels to improve student model robustness, Zhang et al.~\cite{zhang2019theoretically} introduce TRADES to balance robustness and performance on clean examples, and Wang et al.~\cite{wang2019improving} address misclassified examples through MART. Main factors contributing to robustness include larger models, more data, and the use of KL divergence for the inner maximization. Wu et al.'s~\cite{wu2021wider} study confirm this perspective, demonstrating that AT significantly improves robustness in large models, though the improvements were less substantial for smaller models.

\paragraph{Adversarial Robustness Distillation (ARD)} KD does not provide the student model with sufficient robustness, while AT demands a large model capacity. Intuitively, researchers have extensively explored ARD, which enhances student model robustness by combining KD and AT. Revisiting existing ARD methods~\cite{zhu2021reliable,maroto2022benefits}, Robust Soft Label Adversarial Distillation improves robustness by using robust soft labels in inner optimization, highlighting their importance in AT. 
While most methods focus on prediction output, recent research has explored matching feature layers~\cite{shafahi2019adversarially,vaishnavi2022transferring} and input gradients~\cite{chan2020thinks} to generate more robust student models.
In this paper, we conduct an in-depth analysis of mainstream research achievements related to ARD. 
RSLAD~\cite{zi2021revisiting} improves model robustness by introducing robust soft labels, although its accuracy on clean examples still lags behind traditional training. 
Multi-Teacher Adversarial Robust Distillation (MTARD)~\cite{zhao2022enhanced} leverages a dual-teacher framework and adaptive normalized loss function to achieve a better trade-off between student model's robustness and accuracy, yet there remains room for optimization to achieve more stable performance balance. Inspired by these studies, we propose a more effective ARD method to enhance the robustness of student models.

%% file: sec/3_proposed_distillation_strategy.tex
\section{Methodology}
\label{sec:proposed_distillation_strategy}
\subsection{An Adversarial Training Perspective on KD}
As shown in Figure~\ref{fig:illustration}, the traditional KD process focuses on enabling the student model to inherit the teacher model's accuracy without addressing adversarial robustness.
However, traditional KD only allows the student model to inherit the accuracy of natural examples, while its robustness against adversarial examples is significantly lower than that of the teacher model. 
Thus ARD redefines the distillation objective from an adversarial perspective, as follows:\begin{small}
\begin{align}
\mathbb{E}_{p_{d}(x)} \bigg[\underbrace{\alpha \mathcal{L}_{CE}(S(x), y)}_{\text{Clean Example}} + \underbrace{\beta \tau^{2} \mathcal{L}_{KL}\left(S^{\tau}\left(x^{*}\right), T^{\tau}(x)\right)}_{\text{Adversarial Example}}\bigg],
\end{align}
\end{small}where x$^{*}$ denotes the search result of the inner optimization, which can be expressed in the following form:
\begin{equation}
\label{equation_5}
x^{*} = x + \arg\max_{\|\delta\|_{p} \leq \epsilon} \mathcal{L}_{CE}(S(x + \delta), y),
\end{equation}
this constraint ($\|\delta\|_{p} \leq \epsilon$) ensures that the perturbation \( \delta \) is within a specified bound \( \epsilon \) under the \( L_p \)-norm. The \( L_p \)-norm measures the magnitude of the perturbation, and \( \epsilon \) is the maximum allowable perturbation size.

With advancements in ARD research, Zhao et al.~\cite{zhao2022enhanced} design a dynamic training method that can balance the influence of adversarial and non-adversarial teacher models on student models. MTARD extends prior work by incorporating multi-teacher adversarial robustness distillation to guide the adversarial training of lightweight models. The basic min-max optimization framework of MTARD is defined in the following form:
\begin{small}
\begin{align}
\arg\min_{\theta_{S}}\bigg[\underbrace{\alpha \mathcal{L}_{KL}(S(x),T_{nat}(x))}_{\text{Nature Example}} + \underbrace{\beta \mathcal{L}_{KL}\left(S\left(x^{*}\right), T_{adv}(x^{*})\right)}_{\text{Adversarial Example}}\bigg],
\label{eq:mtard}
\end{align}
\end{small}
where $x^{*}$ is an adversarial example generated from the clean example $x$ as shown below:
\begin{align}
x^{*}&=\underset{\delta\in\Omega}{\arg\max}\text{CE}\left(S\left(x+\delta;\theta_{S}\right),y\right),
\label{eq:mtard_sample}
\end{align}
it's worth noting that $\alpha$ and $\beta$ in the above formulas are weighting factors that sum to 1.

\subsection{Framework Overview}
In this section, we give framework details of our proposed CIARD, which simultaneously enhances the robustness of student models while preserving their high accuracy on clean examples. 
As in Figure~\ref{fig:framework}, CIARD achieves high clean accuracy and adversarial robustness by coordinating knowledge transfer between dual teacher models. 
Given a batch of clean examples $x$ $\in$ ${D}_{train}$, these inputs first pass through the clean teacher model $t$ to generate softened class probabilities $t(x)$, serving as reference targets for maintaining natural pattern recognition capabilities. Simultaneously, each clean sample undergoes adversarial perturbation through a Projected Gradient Descent (PGD)~\cite{madry2017towards} attack that jointly considers both the student model $s$ and the clean teacher model $t$:
\begin{small}
\begin{align}
\label{adversarial_sample}
x^* = \text{PGD}_\epsilon(x, s, t) = \underset{\| \delta \|_\infty \leq \epsilon}{\arg\max} \left[ \text{KL}(s(x+\delta) \| t(x+\delta)) \right].
\end{align}
\end{small}

This collaborative attack generation strategy produces adversarial examples $x^{*}$ that challenge both the learning student and the vulnerable clean teacher, ensuring exposure to evolving attack patterns throughout the training process.

At the same time, for each input pair ($x$, $x^*$), the clean teacher processes both the benign examples and adversarial examples simultaneously, generating the corresponding probability distributions $t(x)$ and $t(x^{*})$. Meanwhile, the robust teacher ($t'$) analyzes the adversarial examples at a fixed temperature, producing smoothed outputs $t'(x^{*})$, to maintain its pre-trained defensive decision boundary. The student model processes both types of data through separate forward passes simultaneously, generating $s(x)$ for benign inputs and $s(x^{*})$ for adversarial examples, thereby establishing a dual behavioral baseline for knowledge distillation.
In summary, CIARD's innovation is ultimately manifested in a triple-objective loss function that dynamically balances competing learning goals. The concise knowledge transfer component ($\alpha \text{KL}(s(x), t(x))$) with adaptive temperature initially promotes broad class relationship learning, before gradually sharpening to emphasize discriminative features. Complementarily, the robust knowledge alignment term ($\beta \text{KL}(s(x^{*}), t'(x^{*})$) transplants certified defense mechanisms by enforcing distributional consistency between the student's adversarial responses and the robust teacher's calibrated outputs. 


\subsection{Robust Specialization: Contrastive Push Loss}
\label{subsec:push_loss}
The core challenge in adversarial robustness distillation lies in effectively decoupling clean knowledge from robust knowledge, ensuring that student model can specialize in learning robust features without interference from clean teachers. To address this challenge, we propose ``\textbf{Contrastive Push Loss}'', a novel component designed to enhance the student model's ability to specialize in robustness knowledge.  Formally, given an adversarial sample $x^*$ generated through Eq.~\ref{adversarial_sample}, let $s(x^*)$ and $t(x^*)$ denote the output probability distributions of the student and clean teacher respectively. 
Based on Eq.~\ref{eq:mtard}, for a given input sample $x$ and its adversarial counterpart $x^{*}$, the loss function is formulated as:
\begin{align}
    \label{student_total_loss}
    \mathcal{L}_{student} & = \alpha \underbrace{ \text{KL}(s(x), t(x))}_{\text{Clean Knowledge}} + \beta \underbrace{\text{KL}(s(x^{*}), t'(x^{*}))}_{\text{Robust Knowledge}}\notag \\ & - \lambda \underbrace{\text{Push}(s(x^{*}), t(x^{*}))}_{\text{Robust Specialization}},
\end{align}
where $s$, $t$, and $t'$ represent the student model, clean teacher, and robust teacher respectively.

Unlike conventional distillation that minimizes this divergence, our formulation explicitly maximizes it through negative weighting in the global loss function (see Eq.~\ref{student_total_loss}). This creates a repulsive force that drives the student's adversarial predictions away from the clean teacher's vulnerable patterns while preserving alignment with the robust teacher's guidance through \(\mathcal{L}_{\text{Robust}}\).

As shown in Algorithm~\ref{Push Loss Computation}, our proposed push loss works by deliberately creating a divergence between the student model and the clean teacher. This mechanism allows the student to focus exclusively on robust features when processing adversarial examples, minimizing the influence of potentially misleading robustness knowledge from clean teacher. By pushing away from the clean teacher's incorrect predictions, the student can more effectively absorb the robust teacher's specialized knowledge. Meanwhile, the decoupling process also enhances the clean teacher’s robustness against adversarial data generated by the student model during training. As a result, the remaining unpushed data retains more accurate robust knowledge, which in turn further strengthens the robust training process.
This approach creates a specialized learning environment where the student model can focus on developing robust features without the confusion introduced by clean but non-robust knowledge. By strategically diverging from the clean teacher in cases where robustness is critical, the student achieves better specialization in handling adversarial examples while maintaining performance on clean data.

\subsection{Iterative Teacher Training}
\label{subsec:iterative_teacher}
The cyclic iterative mechanism in CIARD fundamentally re-imagines the teacher-student relationship through the bidirectional flow of knowledge. Unlike traditional distillation methods where teachers stay unchanged, our approach creates a two-way learning process where both the student model improves and the teacher model develops at the same time. As shown in Figure~\ref{fig:framework}, this mechanism operates in the following way:
\begin{align}
    \label{teacher_loss}
    \mathcal{L}_{adv\_teacher} = \text{CE}(t'(x^{*}), y),
\end{align}
where $t'$ represents the robust teacher model, $x^{*}$ denotes the adversarial examples obtained by Eq.~\ref{adversarial_sample}, and $y$ is the ground truth label. 
By optimizing this loss function, the robust teacher model continuously adapts to evolving adversarial attack patterns, maintaining high performance levels throughout the training process. This dynamic adaptation mechanism ensures consistent quality of knowledge transfer and prevents the degradation of the robust teacher's performance that is commonly observed in traditional fixed-teacher approaches.

In our 300-round model training process, we keep the adversarial teacher model fixed for the first 50 rounds, allowing the student model to maximize its learning of the teacher model's robustness. After 50 rounds, as the student model continues training, the generated adversarial examples can cause the teacher model's performance to decline. Therefore, we incorporate the robust teacher model into the training to maintain its performance at a high level.
By optimizing this loss function, the robust teacher model continuously adapts to evolving adversarial examples, maintaining high performance levels throughout the training process. This dynamic adaptation mechanism ensures consistent quality of knowledge transfer and prevents the degradation of the robust teacher's performance that is commonly observed in traditional fixed-teacher approaches.

\begin{algorithm}[t]
\caption{Push Loss Computation}
\label{Push Loss Computation}
\SetAlgoLined
\KwIn{
    $\mathbf{z}_t \in \mathbb{R}^{B \times C}$ -- Teacher logits for a batch of size $B$ and $C$ classes; $\mathbf{z}_s \in \mathbb{R}^{B \times C}$ -- Student logits for the same batch; $\mathbf{y} \in \mathbb{R}^B$ -- Ground truth labels; Temperature $T$ (default: 4).
}
\KwOut{$\mathcal{L}_{\text{push}}$}
\vspace{0.5em}
\textbf{Procedure:} \\
\vspace{0.3em}
\textcolor[rgb]{1,0,0}{$//*$ Clean Teacher Robustness Evaluation $*//$}\\
\quad $\hat{\mathbf{y}}_t \leftarrow \arg\max(\mathbf{z}_t, \text{dim}=1)$ \tcp*{${T}_{predict}$}
\quad $J \leftarrow \{ i \mid \hat{y}_t^{(i)} \neq y^{(i)} \}$ \tcp*{Error indices}
\vspace{0.5em}
\textcolor[rgb]{1,0,0}{$//*$ Clean Teacher Knowledge Filtering $*//$}\\
\quad $\mathbf{z}_t' \leftarrow \{\mathbf{z}_t^{(i)} \mid i \in J\}$\tcp*{Select ${T}_{\text{logits}}$}
\quad $\mathbf{z}_s' \leftarrow \{\mathbf{z}_s^{(i)} \mid i \in J\}$\tcp*{Select ${S}_{\text{logits}}$}
\vspace{0.5em}
\textcolor[rgb]{1,0,0}{$//*$ Robustness Decoupling \& Push Loss $*//$}\\
\quad $\tilde{\mathbf{p}}_s \leftarrow \text{softmax}\left(\frac{\mathbf{z}'_s}{T}\right)$\tcp*{S-Distribution}
\quad $\tilde{\mathbf{p}}_t \leftarrow \text{softmax}\left(\frac{\mathbf{z}'_t}{T}\right)$\tcp*{T-Distribution}
\quad $\mathcal{L}_{\text{push}} \leftarrow D_{\text{KL}}(\tilde{\mathbf{p}}_s \parallel \tilde{\mathbf{p}}_t)$\tcp*{KL divergence}
\textbf{Return} $\mathcal{L}_{\text{push}}$
\end{algorithm}
\subsection{Overall Training Objectives}
To sum up, based on the dual-teacher framework for optimizing the student model, our proposed push loss and iterative teacher training effectively enhance the student model's robustness against adversarial examples while maintaining high performance of the adversarial teacher model throughout the training process, overcoming the performance degradation issues commonly observed in traditional fixed-teacher approaches.

Specifically, for a given dataset $x$, we first obtain its corresponding adversarial examples $x^*$ generated through PGD that jointly consider both the student model $s$ and the clean teacher model $t$. We then input both clean examples and adversarial examples into the clean teacher model $t$, robust teacher model $t'$, and student model $s$ to obtain the corresponding probability distributions $t(x)$, $t(x^*)$, $t'(x^*)$, $s(x)$, and $s(x^*)$, which serve as the foundation for computing the triple-objective loss function, and optimize student model parameters according to the following objective function:
\begin{align}
    \arg \min_{\theta} \mathcal{L}_{student},
\end{align}
where $\alpha$, $\beta$, and $\lambda$ balance clean knowledge transfer, robust knowledge acquisition, and robust specialization, respectively. The detailed algorithm description of CIARD can be found in the supplementary files.


        
        
        

\begin{table*}[htbp]\small
\begin{center}
\setlength{\abovecaptionskip}{0.15cm}
\caption{Performance of Different Teacher Models (ResNet [RN] and WideResNet [WRN]) on CIFAR-10 and CIFAR-100 Datasets.}
\label{tab:tab1}
\vspace{-0.075in}
\setlength{\tabcolsep}{3.2mm}{
\begin{tabular}{c|c|c|c|c|c|c|c}
\toprule
Dataset & Teacher Model & Type & Clean Acc & FGSM~\cite{goodfellow2014explaining} & PGD$_{\textup{SAT}}$~\cite{madry2017towards} & PGD$_{\textup{TRADES}}$~\cite{zhang2019theoretically} & $\textup{CW}_{\infty}$~\cite{carlini2017towards} \\
\midrule
\multirow{2}{*}{CIFAR-10} 
& RN-56     & Clean  &\textbf{93.18\%} & 19.18\% & 0 & 0 & 0 \\
& WRN-34-10 & Robust & 84.92\% & \textbf{60.87\%} & \textbf{56.86\%} & \textbf{55.30\%} &\textbf{53.84\%} \\
\midrule
\multirow{2}{*}{CIFAR-100} 
& WRN-22-6  & Clean  &\textbf{72.55\%} & 25.19\% & 0 & 0 & 0 \\
& WRN-70-16 & Robust & 63.56\% & \textbf{43.69\%} & \textbf{32.24\%} & \textbf{30.95\%} & \textbf{28.93\%} \\
\bottomrule
\end{tabular}
}
\vspace{-0.1in}
\end{center}
\end{table*}

%% file: sec/4_experiments.tex
\section{Experiments}
\label{sec:experiment_setup}
\begin{table}[htbp]
\centering
\small
\caption{White-box Adversarial Robustness of ResNet-18 on CIFAR-10 and CIFAR-100 Datasets. Detailed results from our experiments are presented in the supplementary files. The best results are \textbf{bolded}, and the second best results are \uline{underlined}.}
\label{tab_white_box_results_RN_18}
\setlength{\tabcolsep}{2.0pt}
\resizebox{\linewidth}{!}{
\begin{tabular}{@{}llSS[table-format=2.2]SS[table-format=2.2]SS[table-format=2.2]@{}}
\toprule
\multirow{2}{*}{Attack} & \multirow{2}{*}{Defense} & \multicolumn{3}{c}{CIFAR-10(\%)} & \multicolumn{3}{c@{}}{CIFAR-100(\%)} \\
\cmidrule(lr){3-5} \cmidrule(l){6-8}
 & & {Clean} & {Robust} & {W-R} & {Clean} & {Robust} & {W-R} \\
\midrule
\multirow{10}{*}{FGSM}
& SAT & 84.20 & 55.59 & 69.90 & 56.16 & 25.88 & 41.02 \\
& TRADES & 83.00 & 58.35 & 70.68 & 57.75 & 31.36 & 44.56 \\
& ARD & 84.11 & 58.40 & 71.26 & 60.11 & 33.61 & 46.86 \\
& RSLAD & 83.99 & 60.41 & 72.20 & 58.25 & \uline{34.73} & 46.49 \\
& SCORE & 84.43 & 59.84 & 72.14 & 56.40 & 32.94 & 44.67 \\
& Fair-ARD & 83.41 & 58.91 & 71.16 & 57.81 & 34.39 & 46.10 \\
& ABSLD & 83.21 & 60.22 & 71.72 & 56.77 & \textbf{34.94} & 45.86  \\
& MTARD & 87.36 & 61.20 & 74.28 & 64.30 & 31.49 & 47.90 \\
& B-MTARD & \uline{88.20} & \uline{61.42} & \uline{74.81} & \uline{65.08} & 34.21 & \uline{49.65} \\
& \cellcolor{gray!30}\textbf{CIARD (Ours)} & \textbf{88.87} & \textbf{61.88} & \textbf{75.38} & \textbf{65.73} & 34.47 & \textbf{50.10} \\
\addlinespace
\toprule
\multirow{10}{*}{PGD$_{SAT}$}
& SAT & 84.20 & 45.85 & 65.08 & 56.16 & 21.18 & 38.67 \\
& TRADES & 83.00 & 52.35 & 67.68 & 57.75 & 28.05 & 42.90 \\
& ARD & 84.11 & 50.93 & 67.52 & 60.11 & 29.40 & 44.76 \\
& RSLAD & 83.99 & \uline{53.94} & 68.97 & 58.25 & \uline{31.19} & 44.72 \\
& SCORE & 84.43 & 53.72 & 69.08 & 56.40 & 30.27 & 43.34 \\
& Fair-ARD & 83.41 & 52.00 & 67.71 & 57.81 & 30.64 & 44.23 \\
& ABSLD & 83.21 & \textbf{54.63} & 68.92 & 56.77 & \textbf{32.41} & 44.59 \\
& MTARD & 87.36 & 50.83 & 69.05 & 64.30 & 24.95 & 44.63 \\
& B-MTARD & \uline{88.20} & 51.68 & \uline{69.94} & \uline{65.08} & 28.50 & \uline{46.79} \\
& \cellcolor{gray!30}\textbf{CIARD (Ours)} & \textbf{88.87} & 51.70 & \textbf{70.29} & \textbf{65.73} & 28.05 & \textbf{46.89} \\
\bottomrule
\end{tabular}
}
\footnotesize
\end{table}

\begin{table}[htbp]
\centering
\small
\caption{White-box Adversarial Robustness of MobileNet-V2 on CIFAR-10 and CIFAR-100 Datasets. Detailed results from our experiments are presented in the supplementary files. The best results are \textbf{bolded}, and the second best results are \uline{underlined}.}
\label{tab_white_box_results_MN_V2}
\setlength{\tabcolsep}{2.0pt}
\resizebox{\linewidth}{!}{
\begin{tabular}{@{}llSS[table-format=2.2]SS[table-format=2.2]SS[table-format=2.2]@{}}
\toprule
\multirow{2}{*}{Attack} & \multirow{2}{*}{Defense} & \multicolumn{3}{c}{CIFAR-10(\%)} & \multicolumn{3}{c@{}}{CIFAR-100(\%)} \\
\cmidrule(lr){3-5} \cmidrule(l){6-8}
 & & {Clean} & {Robust} & {W-R} & {Clean} & {Robust} & {W-R} \\
\midrule
\multirow{10}{*}{FGSM}
& SAT & 83.87 & 55.89 & 69.88 & 59.19 & 30.88 & 45.04 \\
& TRADES & 77.95 & 53.75 & 65.85 & 55.41 & 30.28 & 42.85 \\
& ARD & 83.43 & 57.03 & 70.23 & 60.45 & 32.77 & 46.61 \\
& RSLAD & 83.20 & \textbf{59.47} & 71.34 & 59.01 & 33.88 & 46.45 \\
& SCORE & 82.32 & 58.43 & 70.38 & 49.38 & 29.28 & 39.33 \\
& Fair-ARD & 82.65 & 56.37 & 69.51 & 59.18 & \uline{34.07} & 46.63 \\
& ABSLD & 82.50 & 58.47 & 70.49 & 56.67 & 33.85 & 45.26 \\
& MTARD & \uline{89.26} & 57.84 & 73.55 & \textbf{67.01} & 32.42 & 49.72 \\
& B-MTARD & 89.09 & 58.79 & \uline{73.94} & 66.13 & \textbf{34.36} & \textbf{50.25} \\
& \cellcolor{gray!30}\textbf{CIARD (Ours)} & \textbf{89.51} & \uline{59.10} & \textbf{74.31} & \uline{66.72} & 33.56 & \uline{50.14} \\
\addlinespace
\toprule
\multirow{10}{*}{PGD$_{SAT}$}
& SAT & 83.87 & 46.84 & 65.36 & 59.19 & 25.64 & 42.42 \\
& TRADES & 77.95 & 49.06 & 63.51 & 55.41 & 23.33 & 39.37 \\
& ARD & 83.43 & 49.50 & 66.47 & 60.45 & 28.69 & 44.57 \\
& RSLAD & 83.20 & \uline{53.25} & 68.23 & 59.01 & \uline{30.19} & 44.60 \\
& SCORE & 82.32 & \textbf{53.42} & 67.87 & 49.38 & 27.03 & 38.21 \\
& Fair-ARD & 82.65 & 50.50 & 66.58 & 59.18 & 30.15 & 44.67 \\
& ABSLD & 82.50 & 52.98 & 67.74 & 56.67 & \textbf{31.28} & 43.98 \\
& MTARD & \uline{89.26} & 44.16 & 66.71 & \textbf{67.01} & 25.14 & 46.08 \\
& B-MTARD & 89.09 & 47.56 & \uline{68.33} & 66.13 & 28.47 & \textbf{47.30} \\
& \cellcolor{gray!30}\textbf{CIARD (Ours)} & \textbf{89.51} & 47.67 & \textbf{68.59} & \uline{66.72} & 27.02 & \uline{46.87} \\
\bottomrule
\end{tabular}
}
\footnotesize
\end{table}
\begin{table}[t!]\small
\centering
\small
\caption{Black-box Adversarial Robustness of ResNet-18 on CIFAR-10 and CIFAR-100 Datasets Under Various Attack Methods. Performance is measured using Weighted Robustness (W-R) and accuracy metrics. Additional experimental details are provided in the supplementary materials.}
\label{tab_black_box_cifar10_cifar100_ResNet-18}
\setlength{\tabcolsep}{1.5pt}
\resizebox{\linewidth}{!}{
\begin{tabular}{@{}llSS[table-format=2.2]SS[table-format=2.2]SS[table-format=2.2]@{}}
\toprule
\multirow{2}{*}{Attack} & \multirow{2}{*}{Defense} & \multicolumn{3}{c}{CIFAR-10(\%)} & \multicolumn{3}{c@{}}{CIFAR-100(\%)} \\
\cmidrule(lr){3-5} \cmidrule(l){6-8}
 & & {Clean} & {Robust} & {W-R} & {Clean} & {Robust} & {W-R} \\
\midrule
\multirow{7}{*}{PGD\textsubscript{TRADES}}
& SAT        & 84.20  & 64.74  & 74.52 & 56.16  & 38.10  & 47.13 \\
& TRADES     & 83.00  & 63.61  & 73.31 & 57.75  & 38.20  & 47.98 \\
& ARD        & 84.11  & 63.50  & 73.81 & 60.11  & 39.53  & 49.82 \\
& RSLAD      & 83.99  & 63.96  & 73.98 & 58.25  & 39.93  & 49.09 \\
& MTARD      & 87.36  & 65.26  & 76.31 & 64.30  & 41.46  & 52.88 \\
& B-MTARD    & \uline{88.20}  & \uline{65.29}  & \uline{76.75} & \uline{65.08}  & \uline{42.11}  & \uline{53.60} \\
& \cellcolor{gray!30}\textbf{CIARD (Ours)} & \textbf{88.87} & \textbf{66.28} & \textbf{77.58} & \textbf{65.73} & \textbf{42.29} & \textbf{54.01} \\
\addlinespace
\toprule
\multirow{7}{*}{CW$_{\infty}$}
& SAT        & 84.20  & 63.84  & 74.02 & 56.16  & 39.42  & 47.79 \\
& TRADES     & 83.00  & 62.83  & 72.92 & 57.75  & 38.63  & 48.19 \\
& ARD        & 84.11  & 62.86  & 73.49 & 60.11  & 38.85  & 49.48 \\
& RSLAD      & 83.99  & 63.05  & 73.52 & 58.25  & 39.67  & 48.96 \\
& MTARD      & 87.36  & 64.58  & 75.97 & 64.30  & 41.18  & 52.74 \\
& B-MTARD    & \uline{88.20}  & \uline{64.64}  & \uline{76.42} & \uline{65.08}  & \uline{41.35}  & \uline{53.22} \\
& \cellcolor{gray!30}\textbf{CIARD (Ours)} & \textbf{88.87} & \textbf{64.79} & \textbf{76.83} & \textbf{65.73} & \textbf{41.44} & \textbf{53.59} \\
\addlinespace
\toprule
\multirow{7}{*}{SA~\cite{andriushchenko2020square}}
& SAT        & 84.20  & 72.48  & 78.34 & 56.16  & 40.05  & 48.11 \\
& TRADES     & 83.00  & 72.49  & 77.75 & 57.75  & 42.51  & 50.13 \\
& ARD        & 84.11  & 74.60  & 79.36 & 60.11  & 47.20  & 53.66 \\
& RSLAD      & 83.99  & 72.47  & 78.23 & 58.25  & 45.32  & 51.79 \\
& MTARD      & 87.36  & 78.58  & 82.97 & 64.30  & 48.13  & 56.22 \\
& B-MTARD    & \uline{88.20}  & \uline{79.82}  & \uline{84.01} & \uline{65.08}  & \uline{49.40}  & \uline{57.24} \\
& \cellcolor{gray!30}\textbf{CIARD (Ours)} & \textbf{88.87} & \textbf{80.03} & \textbf{84.45} & \textbf{65.73} & \textbf{49.76} & \textbf{57.75} \\
\bottomrule
\end{tabular}
}
\footnotesize
\end{table}

\begin{table}[t!]\small
\centering
\small
\caption{Black-box Adversarial Robustness of MobileNet-V2 on CIFAR-10 and CIFAR-100 Datasets Under Various Attack Methods. Performance is measured using Weighted Robustness (W-R) and accuracy metrics. Additional experimental details are provided in the supplementary materials.}
\label{tab_black_box_cifar10_cifar100_MobileNet-V2}
\setlength{\tabcolsep}{1.5pt}
\resizebox{\linewidth}{!}{
\begin{tabular}{@{}llSS[table-format=2.2]SS[table-format=2.2]SS[table-format=2.2]@{}}
\toprule
\multirow{2}{*}{Attack} & \multirow{2}{*}{Defense} & \multicolumn{3}{c}{CIFAR-10(\%)} & \multicolumn{3}{c@{}}{CIFAR-100(\%)} \\
\cmidrule(lr){3-5} \cmidrule(l){6-8}
 & & {Clean} & {Robust} & {W-R} & {Clean} & {Robust} & {W-R} \\
\midrule
\multirow{7}{*}{PGD\textsubscript{TRADES}}
& SAT        & 83.87  & 64.66  & 74.27                     &59.19 &40.70 &49.95 \\
& TRADES     & 77.95  & 61.04  & 69.50                     &55.41 &37.76 &46.59 \\
& ARD        & 83.43  & 63.28  & 73.36                     &60.45 &39.15 &49.80 \\
& RSLAD      & 83.20  & 64.33  & 73.77                     &59.01 &40.32 &49.67 \\
& MTARD      & \uline{89.26}  & 66.30  & 77.78             &\textbf{67.01} &\textbf{43.23} &\textbf{55.12}  \\
& B-MTARD    & 89.09  & \uline{66.47}  & \uline{77.78}     &66.13 &42.67 &54.40  \\
& \cellcolor{gray!30}\textbf{CIARD (Ours)}      & \textbf{89.51} & \textbf{66.66}  & \textbf{78.09}   & \uline{66.72}  & \uline{42.70}  & \uline{54.71} \\
\addlinespace
\toprule
\multirow{7}{*}{CW$_{\infty}$}
& SAT        & 83.87  & 64.24  & 74.06                     &59.19  &40.97 &50.08 \\
& TRADES     & 77.95  & 60.66  & 69.31                     &55.41  &38.02 &46.72 \\
& ARD        & 83.43  & 62.83  & 73.13                     &60.45  &38.53 &49.49 \\
& RSLAD      & 83.20  & 63.45  & 73.33                     &59.01  &39.92 &49.47 \\
& MTARD      & \uline{89.26}  & 65.68  & 77.47             &\textbf{67.01}  &\textbf{42.92} &\textbf{54.97} \\
& B-MTARD    & 89.09  & \uline{65.96}  & \uline{77.53}     &66.13  &42.04 &54.09 \\
& \cellcolor{gray!30}\textbf{CIARD (Ours)}     & \textbf{89.51} & \textbf{66.12}  & \textbf{77.82}   & \uline{66.72}  & \uline{42.85}  & \uline{54.79} \\
\addlinespace
\toprule
\multirow{7}{*}{SA~\cite{andriushchenko2020square}}
& SAT        & 83.87  & 73.01  & 78.44                     &59.19   &44.51 &51.85  \\
& TRADES     & 77.95  & 67.43  & 62.69                     &55.41   &40.71 &48.06 \\
& ARD        & 83.43  & 73.26  & 78.35                     &60.45   &46.95 &53.70 \\
& RSLAD      & 83.20  & 73.07  & 78.14                     &59.01   &45.66 &52.34  \\
& MTARD      & \uline{89.26}  & 79.13  & 84.20             &\textbf{67.01}   &50.64 &\textbf{58.83}  \\
& B-MTARD    & 89.09  & \uline{79.61}  & \uline{84.35}     &66.13   &\uline{50.83} &58.48  \\
& \cellcolor{gray!30}\textbf{CIARD (Ours)}      & \textbf{89.51} & \textbf{80.01}  & \textbf{84.76}   & \uline{66.72}  & \textbf{50.85}  & \uline{58.79} \\
\bottomrule
\end{tabular}
}
\footnotesize
\end{table}

\subsection{Experimental Setup}
\noindent\textbf{Datasets \& Models.} We use the CIFAR-10~\cite{krizhevsky2009learning} and CIFAR-100 datasets, following Zhao et al.~\cite{zhao2022enhanced}, to evaluate the experimental results of CIARD and other ARD methods. For model architecture, the student model uses ResNet-18~\cite{he2016deep} and MobileNet-V2~\cite{sandler2018mobilenetv2}. The teacher models are categorized into a clean teacher and an adversarial teacher. The clean teacher uses ResNet-56 for CIFAR-10 and WideResNet-22-6~\cite{zagoruyko2016wide} for CIFAR-100. The adversarial teacher uses WideResNet-34-10 trained with TRADES~\cite{zhang2019theoretically} for CIFAR-10 and WideResNet-70-16 provided by Gowal et al~\cite{gowal2020uncovering}, for CIFAR-100. The performance of these pre-trained teacher models is summarized in Table~\ref{tab:tab1}. 

\noindent\textbf{Implementation Details.}
Student models are trained for 300 epochs using an SGD optimizer (momentum 0.9, weight decay 2e-4), with a learning rate following a cosine decay schedule from 0.1 to 1e-5. The adversarial teacher is frozen for the first 50 epochs and then iteratively updated using SGD with a low learning rate of 1e-5. For robust training, adversarial examples are generated via a 10-step PGD attack with a step size of 2/255 and an $L_{\infty}$ bound of $\epsilon=8/255$. The push loss temperature is set to 4. All experiments use a batch size of 64 and standard data augmentation (random cropping and horizontal flipping). Training was conducted in PyTorch on two NVIDIA RTX 4090 GPUs.

\noindent\textbf{Evaluation Metrics.}
To evaluate model performance, we measure natural accuracy on clean test examples and robust accuracy on adversarial test examples. The evaluation protocol encompasses multiple attack methods: FGSM~\cite{goodfellow2014explaining}, PGD$_{\textup{SAT}}$~\cite{madry2017towards}, PGD$_{\textup{TRADES}}$~\cite{zhang2019theoretically}, $\textup{CW}_{\infty}$~\cite{carlini2017towards}, and Square Attack~\cite{andriushchenko2020square}, all configured with a maximum perturbation limit of $\epsilon$ = 8/255. Both PGD$_{\textup{SAT}}$ and PGD$_{\textup{TRADES}}$ are implemented with 20 steps and a step size of 2/255, while $\textup{CW}_{\infty}$ employs 30 steps. For the query-based attack, we set the number of queries for the Square Attack to 100.

\subsection{Effectiveness of CIARD}
Adversarial attacks can be categorized into two types based on the threat model: white-box and black-box attacks. In a white-box setting, the attacker has complete access to the deep learning model, including its architecture and parameters. In contrast, a black-box setup only allows the attacker to access the model's output. Due to the length of the article, more detailed experimental results can be found in the supplementary files.

\noindent\textbf{White-box Robustness of Student Models.}
\label{Section:White-Box_Robustness}
To evaluate robustness against white-box attacks, we test the student models ResNet-18 and MobileNet-V2 on CIFAR-10 and CIFAR-100 using four attack methods: FGSM, PGD$_{\textup{SAT}}$, PGD$_{\textup{TRADES}}$, and $\textup{CW}_{\infty}$.
As shown in Tables~\ref{tab_white_box_results_RN_18} and ~\ref{tab_white_box_results_MN_V2}, the precision of CIARD W-Robust on CIFAR-10 and CIFAR-100 outperforms other ARD methods in most cases. For ResNet-18, the model weight robustness is improved by up to 0.57\% and 0.45\% based on the CIFAR-10 and CIFAR-100 datasets. Furthermore, the cyclic iteration mechanism in CIARD improves the accuracy of the student model's classification, reaching 88.87\% and 65.73\% in CIFAR-10 and CIFAR-100, respectively. For MobileNet-V2, the results are similar, particularly under FGSM attack. Tests based on the CIFAR-10 dataset show an improvement in weighted robustness of 0.37\% and 0.26\% compared to the best benchmark method.

\noindent\textbf{Black-box Robustness of Student Models.}
We also perform black-box evaluations, incorporating both transfer-based and query-based~\cite{andriushchenko2020square} methods, to assess the student model's robustness in environments more akin to real-world scenarios. The parameter configurations for the student and teacher models are consistent with those used in the white-box evaluations. 

Using CIFAR-10 and CIFAR-100, we evaluate the defensive capabilities of CIARD and other methods against black-box attacks on ResNet-18 and MobileNet-V2, focusing on both transfer-based and query-based attacks. For transfer-based attacks, we use adversarial teachers (WideResNet-34-10 and WideResNet-70-16) to generate adversarial examples for PGD$_\text{TRADES}$ and CW$_{\infty}$ attacks. For query-based attacks, we employe Square Attack (SA). We select the best checkpoints for the baseline model and MTARD based on weighted robustness accuracy. Tables~\ref{tab_black_box_cifar10_cifar100_ResNet-18} and ~\ref{tab_black_box_cifar10_cifar100_MobileNet-V2} indicate that CIARD generally exhibits stronger resilience against the three types of black-box attacks. Notably, against PGD$_\text{TRADES}$ attacks, the student model achieve weighted robustness accuracy in CIFAR-10 that was 0.83\% and 0.41\% higher than the second-best methods, respectively. In summary, the experimental results clearly demonstrate the superior performance of CIARD in defending against both white-box and black-box attacks. Based on experimental results, ARD, RSLAD, MTARD, and CIARD outperform SAT and TRADES, demonstrating that ARD methods are more effective in enhancing lightweight model performance compared to traditional approaches.

\begin{figure}[t!]
    \centering
    \includegraphics[width=1\linewidth]{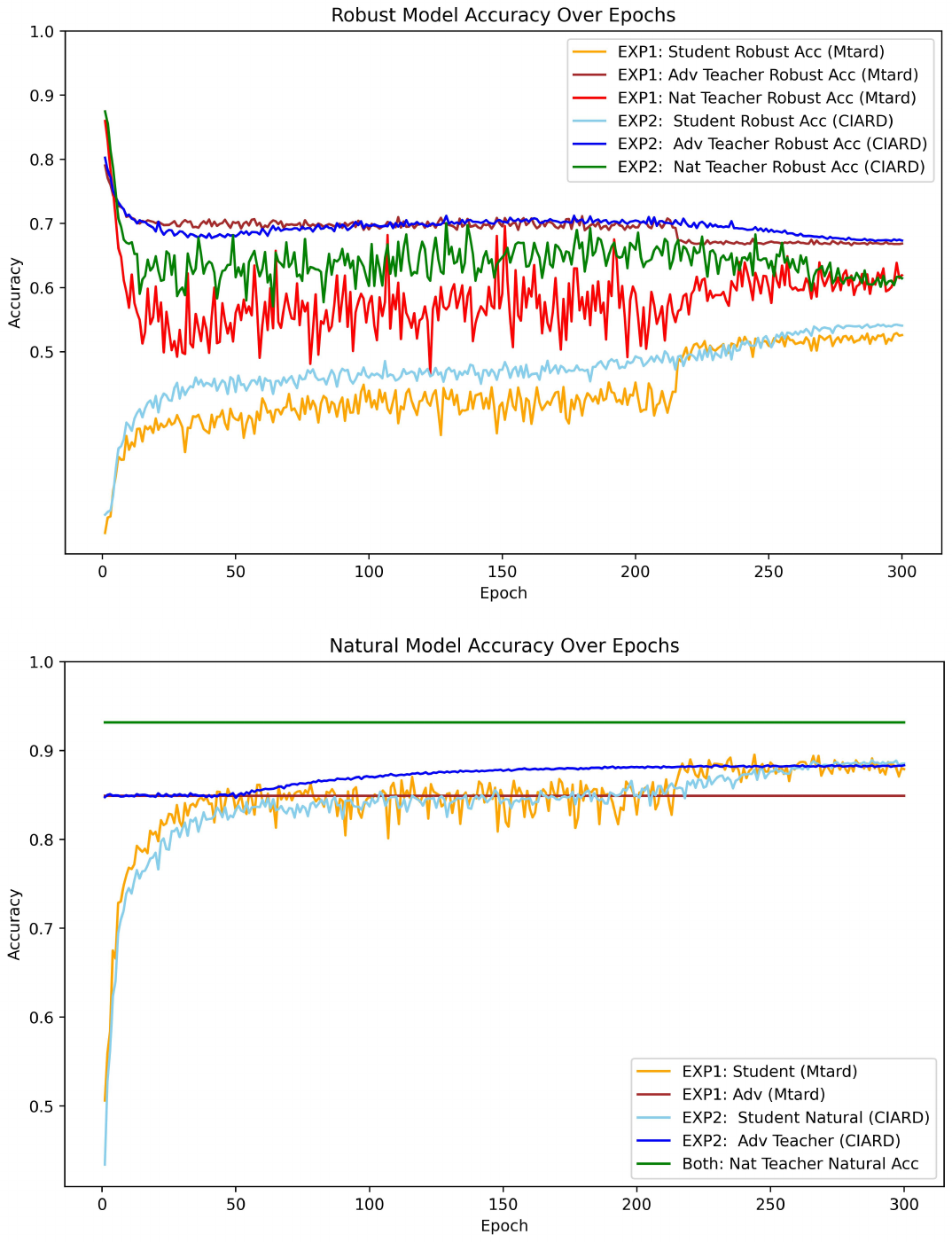}
    \caption{Training curves of ResNet-18 on CIFAR-10 over 300 epochs. Up: Robust accuracy comparison of MTARD/CIARD. Down: Classification accuracy comparison of MTARD/CIARD.}
    \label{fig:Robust_Clean}
\end{figure}
\subsection{Ablation Studies} 
\noindent \textbf{Quantitative Analysis of CIARD}.
To better understand the contribution of each component in our proposed method, we conduct ablation studies on both Iterative Teacher Training (ITT) and Push loss. Table~\ref{tab:cifar10_results} presents the performance comparison on PGD adversarial accuracy (PGD\textsubscript{TRADES}), clean accuracy (Clean\textsubscript{acc}), and weighted accuracy (W-acc). 

\noindent \textbf{The effects of ITT}. From the results, we observe that incorporating ITT improves both clean accuracy and adversarial robustness compared to the baseline model (without ITT and Push loss). Specifically, ITT alone increases clean accuracy by 0.87\% and adversarial robustness by 0.19\%. This improvement demonstrates the effectiveness of our implicit teacher training strategy in enhancing model generalization.

\noindent \textbf{The effects of contrastive push loss}. We further incorporate the push loss into our framework, we achieve the best performance across all metrics. The full model achieves 54.54\% on PGD\textsubscript{TRADES}, 88.86\% on Clean\textsubscript{acc}, and 71.70\% on W-acc, showing consistent improvements over both the baseline and the model with only ITT. The Push loss contributes an additional 0.51\% gain in adversarial robustness and 0.23\% in clean accuracy.
The effectiveness of our contrastive push loss can be attributed to its decoupling design, which helps separate the decision boundaries between clean and adversarial examples. As shown in our prior analysis, this decoupling mechanism allows the clean teacher to maintain higher robustness throughout the training process. Meanwhile, the robust teacher's capability continuously improves as training progresses, benefiting from the knowledge transfer facilitated by the push loss.

\noindent \textbf{Qualitative Analysis of CIARD}.
In addition to the above experimental results, line chart~\ref{fig:Robust_Clean} also validates our design choices and confirms that ITT and Push loss are both key components of our framework, each contributing to overall performance in terms of robustness and accuracy. Despite the clean teacher being frozen to preserve its high clean accuracy, its robust accuracy has significantly improved during training, demonstrating the effectiveness of our decoupling mechanism with push loss, as mentioned in \ref{subsec:push_loss}. Moreover, both the robust and clean accuracy of our robust teacher have improved and remained consistently high, further validating the contribution of our Iterative Teacher Training method.

\begin{table}[t!]
\centering
\caption{ResNet-18 Results on CIFAR-10. Push Loss refers to Contrastive Push Loss in \ref{subsec:push_loss}, while ITT represents the Iterative Teacher Training in \ref{subsec:iterative_teacher}.}
\label{tab:cifar10_results}
\setlength{\tabcolsep}{12pt}
\resizebox{\linewidth}{!}{
\begin{tabular}{@{} l SSS[table-format=1.3] @{}}
\toprule
\textbf{Method} & \textbf{PGD\textsubscript{TRADES}} & \textbf{Clean\textsubscript{acc}} & \textbf{W-acc} \\
\midrule
w/o ITT  w/o Push loss & 53.84 & 87.76 & 70.80 \\ \midrule
w/  ITT  w/o Push loss& 54.03 & 88.63 & 71.30 \\ \midrule
w/  ITT  w/  Push loss& 54.54 & 88.86 & 71.70 \\
\bottomrule
\end{tabular}
}
\vspace{-0.5em}
\footnotesize
\end{table}

%% file: sec/6_conclusion.tex
\section{Conclusion}
\label{sec:conclusion}
In this paper, we propose the Cyclic Iterative Adversarial Robustness Distillation (CIARD) method to address the conflict between the teacher optimization objective and performance degradation in adversarial robustness distillation. Our framework introduces: \textcircled{1} a contrastive push-pull alignment mechanism to resolve the objective conflict by regulating the distance between the student adversarial response and the teacher clean features; and \textcircled{2} dynamic adversarial retraining to preserve the teacher robustness via parameter freezing and adaptive knowledge adjustment. Extensive experiments on CIFAR-10/100 and Tiny-ImageNet demonstrate the effectiveness of CIARD. The method sets a new benchmark for compact adversarial models and is planned to be extended to various attack scenarios and resource-constrained deployment in the future.

\section{Acknowledgment}
Liming Lu, Xiang Gu and Shuchao Pang are supported by the National Natural Science Foundation of China (Grant No.62206128), National Key Research and Development Program of China under (Grant No.2023YFB2703900) and the Postgraduate Research \& Practice Innovation Program of Jiangsu Province (Grant No.KYCX24\_0723).
Anan Du is supported by the Start-up Fund for New Talented Researchers
of Nanjing University of Industry Technology (Grant No.YK24-05-04)
Yongbin Zhou is supported by the National Natural Science Foundation of China (Grant No.U2336205).